\documentclass[a4paper]{article}
\usepackage{fullpage}
\usepackage{authblk}

\usepackage[utf8]{inputenc} 
\usepackage[T1]{fontenc}    
\usepackage{hyperref}   
\usepackage{url}            
\usepackage{booktabs}      
\usepackage{amsfonts}      
\usepackage{nicefrac}      
\usepackage{microtype}     

\usepackage{graphicx} 
\usepackage{bm,dsfont}
\usepackage{amssymb,amsmath,amsthm,mathtools}
\usepackage{multirow}
\usepackage{threeparttable}
\usepackage{rotating}
\usepackage{tablefootnote}
\usepackage{sidecap}
\usepackage{subcaption}
\usepackage{wrapfig}
\usepackage{enumitem}
\usepackage{pdfpages}

\theoremstyle{definition}
\newtheorem{defn}{Definition}
\newtheorem{thm}{Theorem}
\newcommand{\tr}{\text{tr}}

\sidecaptionvpos{figure}{c}

\title{Towards Certified Robustness of Distance Metric Learning}

\author[1]{Xiaochen Yang\footnote{Equal contribution}\footnote{Corresponding author; Email address: xiaochen.yang@glasgow.ac.uk}}
\author[2]{Yiwen Guo$^\ast$}
\author[3]{Mingzhi Dong}
\author[4]{Jing-Hao Xue}
\affil[1]{School of Mathematics and Statistics, Glasgow, UK.}
\affil[3]{School of Computer Science, Fudan University, and Shanghai Key Laboratory of Data Science, Fudan University, Shanghai, China.}
\affil[4]{Department of Statistical Science, University College London, UK.}
\date{}

\begin{document}

\maketitle

\begin{abstract}
Metric learning aims to learn a distance metric such that semantically similar instances are pulled together while dissimilar instances are pushed away. Many existing methods consider maximizing or at least constraining a distance margin in the \emph{feature} space that separates similar and dissimilar pairs of instances to guarantee their generalization ability. In this paper, we advocate imposing an adversarial margin in the \emph{input} space so as to improve the generalization and robustness of metric learning algorithms. We first show that, the adversarial margin, defined as the distance between training instances and their closest adversarial examples in the input space, takes account of both the distance margin in the feature space and the correlation between the metric and triplet constraints. Next, to enhance robustness to instance perturbation, we propose to enlarge the adversarial margin through minimizing a derived novel loss function termed the perturbation loss. The proposed loss can be viewed as a data-dependent regularizer and easily plugged into any existing metric learning methods. Finally, we show that the enlarged margin is beneficial to the generalization ability by using the theoretical technique of algorithmic robustness. Experimental results on 16 datasets demonstrate the superiority of the proposed method over existing state-of-the-art methods in both discrimination accuracy and robustness against possible noise.
\end{abstract}

\section{Introduction}

Metric learning focuses on learning similarity or dissimilarity between data. Research on metric learning originates from at least 2002, where~\cite{xing2002distance} first proposes to formulate it as an optimization problem. Since then, many metric learning methods have been proposed for classification~\cite{kulis2013metric,bellet2015metric,li2018survey}, clustering~\cite{xiang2008learning}, and information retrieval~\cite{mcfee2010metric,yang2018retrieving}. In particular, the methods have shown to be particularly superior in open-set classification and few-shot classification with notable applications in, for example, face verification~\cite{koestinger2012large,hu2017local} and person re-identification~\cite{ma2014person,zou2021person}. 

One commonly studied distance metric is the generalized Mahalanobis distance, which defines the distance between any two instances $\bm x_i, \bm x_j \in \mathbb{R}^p$ as 
\[d_{\bm M}(\bm x_i, \bm x_j) = \sqrt{(\bm x_i - \bm x_j)^T \bm M (\bm x_i - \bm x_j)} ,\]
where $\bm M$ is a positive semidefinite~(PSD) matrix. Owing to its PSD property, $\bm M$ can be decomposed into $\bm L^T \bm L$. Thus, computing the Mahalanobis distance is equivalent to linearly transforming the instances from the input space to the feature space via $\bm L$ and then computing the Euclidean distance $\|\bm L \bm x_i - \bm L \bm x_j \|_2$ in the transformed space. 

To learn a specific distance metric for each task, prior knowledge on instance similarity and dissimilarity should be provided as side information. Metric learning methods differ by the form of side information they use and the supervision encoded in similar and dissimilar pairs. For example, pairwise constraints enforce the distance between instances of the same class to be small (or smaller than a threshold value) and the distance between instances of different classes to be large (or larger than a threshold value)~\cite{xing2002distance,xiang2008learning}. The thresholds could be either pre-defined or learned for similar and dissimilar pairs~\cite{davis2007information,kwok2003learning}. In triplet constraints $(\bm x_i, \bm x_j, \bm x_l)$, distance between the different-class pair $(\bm x_i, \bm x_l)$ should be larger than distance between the same-class pair $(\bm x_i, \bm x_j)$, and typically, plus a margin~\cite{weinberger2009distance,shi2014sparse,song2017parameter,dong2019learning}. More recently, quadruplet constraints are proposed, which require the difference in the distance of two pairs of instances to exceed a margin~\cite{law2013quadruplet}, and $(N+1)$-tuplet extends the triplet constraint for multi-class classification~\cite{sohn2016improved}.

\begin{figure}[!t]
    \centering
    \includegraphics[width=.8\linewidth]{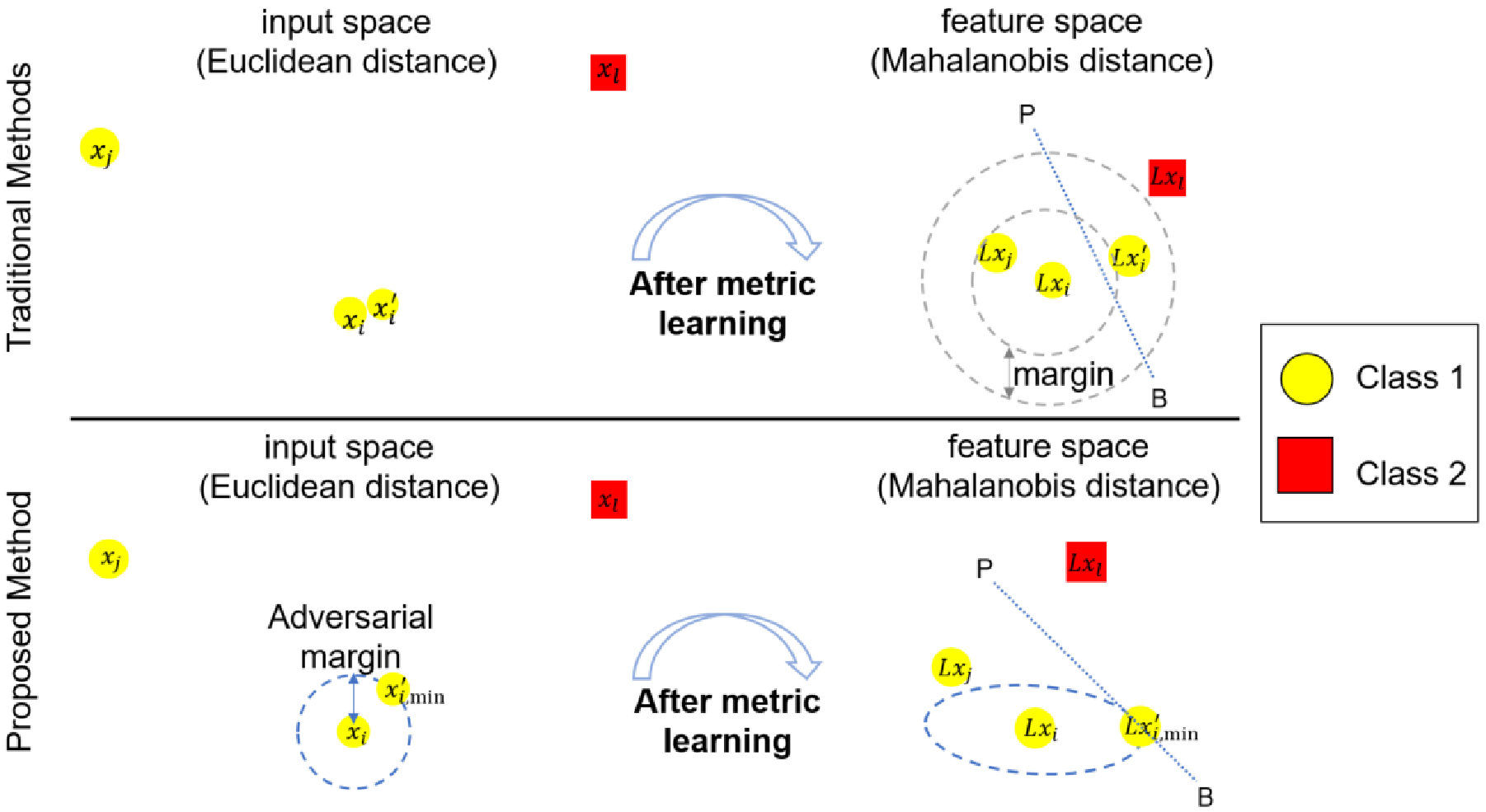}
    \caption{\textit{Upper}: Traditional methods aim to separate the same-class pair $(\bm x_i, \bm x_j)$ and the different-class pair $(\bm x_i, \bm x_l)$ by a margin in the feature space. While $\bm x_i$ has $\bm x_j$ as its nearest neighbor~(NN) in the feature space and is correctly predicted by using the NN classifier, the metric is sensitive to perturbation in the input space; a tiny perturbation from $\bm x_i$ to $\bm x’_i$ changes the NN to $\bm x_l$ and leads to an incorrect prediction. 
    \textit{Bottom}: The proposed method aims to enlarge the adversarial margin in the input space, which equals to the Euclidean distance between $\bm x_i$ and the closest point $\bm x_{i,{\rm min}}$ in the input space that lies on the decision boundary in the feature space (indicated by $PB$) and quantifies the maximum degree to which robustness can be certified.}
    \label{fig:motivation}
\end{figure}

The gap between thresholds in pairwise constraints and the margin in triplet and quadruplet constraints are both designed to learn a distance metric that could ensure good generalization of the subsequent $k$-nearest neighbor~($k$NN) classifier. However, such a distance margin imposed in the feature space does not consider the correlation between the data and the learned metric. Consequently, it may be insufficient to withstand a small perturbation of the instance occurred in the input space, thereby failing to certify the robustness or even possess the anticipated generalization benefit. As illustrated in Fig.~\ref{fig:motivation}(upper), while $\bm x_i$ selects the same-class instance $\bm x_j$ as its nearest neighbor in the feature space, a tiny perturbation from $\bm x_i$ to $\bm x'_i$ in the input space can be magnified by the learned distance metric, leading to a change in its nearest neighbor from $\bm x_j$ to the different-class instance $\bm x_l$. When the NN algorithm is used as the classifier, the perturbation results in an incorrect label prediction. 

In this paper, we propose a simple yet effective method to enhance the generalization ability of metric learning algorithms and their robustness against instance perturbation. As shown in Fig.~\ref{fig:motivation}(bottom), the principal idea is to enlarge the adversarial margin, defined as the distance between a training instance and its closest adversarial example in the input space~\cite{szegedy2013intriguing}. 

In particular, our contributions are fourfold.
\begin{enumerate}
    \item We identify that the distance margin, widely used in existing methods, is insufficient to withstand adversarial examples, and we introduce a direct measure of robustness termed the adversarial margin, which quantifies the maximum degree to which a training instance could be perturbed without changing the label of its nearest neighbor (or $k$ nearest neighbors if required) in the feature space. Building on a geometric insight, we derive an analytically simple solution to the adversarial margin, which reveals the importance of an adaptive margin considering the correlation between the data and the distance metric (Section~\ref{subsec:motivation}-\ref{subsec:support_point}). 
    \item We define a novel hinge-like perturbation loss to penalize the adversarial margin for being small. The proposed loss function serves as a general approach to enhancing robustness, as it can be optimized jointly with any existing triplet-based metric learning methods; the optimization problem suggests that our method learns a discriminative metric in a weighted manner and simultaneously functions as a data-dependent regularization (Section~\ref{subsec:optimization}).
    \item We show the benefit of enlarging the adversarial margin to the generalization ability of the learned distance metric by using the theoretical technique of algorithmic robustness~\cite{xu2012robustness} (Theorem~\ref{thm:bound}, Section~\ref{subsec:bound}). 
    \item We conduct experiments on 16 datasets in both noise-free and noisy settings. Results show that the proposed method outperforms state-of-the-art robust metric learning methods in terms of classification accuracy and validate its robustness to possible noise in the input space (Section~\ref{sec:exp}).
\end{enumerate}

\subsubsection*{Notation}
Let $\{\bm x_i, y_i \}_{i=1}^n$ denote the set of training instance and label pairs, where $\bm x_i \in \mathcal{X} \subseteq \mathbb{R}^p$ and $y_i \in \mathcal{Y} =  \{1, \ldots, C\}$; $\mathcal{X}$ is called the input space. Our framework is based on triplet constraints $\{\bm x_i, \bm x_j, \bm x_l\}$ and we adopt the following strategy for generating triplets~\cite{weinberger2009distance}:
\begin{equation*}
\begin{split}
    \mathcal{S}&= \big\{(\bm x_i, \bm x_j):\bm x_j \in \{k\text{NNs with the same class label of }  \bm x_i\}\big\}, \\
    \mathcal{R}&= \big\{(\bm x_i, \bm x_j, \bm x_l):(\bm x_i, \bm x_j) \in \mathcal{S},y_i \neq y_l \big\}.
\end{split} 
\end{equation*}
$\bm x_j$ is termed the target neighbor of $\bm x_i$ and $\bm x_l$ is termed the impostor. $|\mathcal{S}|$ and $|\mathcal{R}|$ denote the numbers of elements in the sets $\mathcal{S}$ and $\mathcal{R}$, respectively. $d_{\bm E}$ and $d_{\bm M}$ denote the Euclidean and Mahalanobis distances, respectively; $\bm M \in \mathbb{S}^p_+$, where $\mathbb{S}^p_+$ is the cone of $p \times p$ real-valued PSD matrices. $\bm M^2 = \bm M \bm M$. $\mathds{1}[\cdot]$ denotes the indicator function and $[a]_+=\max(a,0)$ for $a\in \mathbb R$.

\section{Methodology}
\label{sec:method}

In this section, we introduce our method for enhancing robustness of triplet-based metric learning algorithms through maximizing the adversarial margin. First, we review the existing distance margin and provide the rationale for enlarging the adversarial margin. Second, an explicit formula for the adversarial margin is derived. Third, we propose the perturbation loss to encourage a larger adversarial margin and present its optimization jointly with the existing large (distance) margin nearest neighbor~(LMNN) algorithm. Lastly, we show that enlarging the adversarial margin is beneficial to the generalization ability of the learned distance metric.

\subsection{Motivation for enlarging the adversarial margin}
\label{subsec:motivation}

Suppose $\bm x_i$ is a training instance and $\bm x_j$, $\bm x_l$ are the nearest neighbor of $\bm x_i$ from the same class and from the different class respectively. Many triplet-based methods, such as LMNN~\cite{weinberger2009distance}, impose the following constraint on the triplet:
\[f(\bm x_i) \coloneqq d^2_{\bm M}(\bm x_i, \bm x_l) - d^2_{\bm M}(\bm x_i, \bm x_j) \geq 1.\]
When the constraint is satisfied, $\bm x_i$ will be correctly classified using the NN classifier. Moreover, the value one represents the unit margin at the distance level and is designed to robustify the model against small noises in training instances. 

Nevertheless, the distance margin may be insufficient to withstand deliberately manipulated perturbations. Let $\Delta \bm x_i$ denote a perturbation of $\bm x_i$. When the perturbation size is constrained as $\|\Delta \bm x_i\|_2 \leq r$, $f(\bm x_i + \Delta \bm x_i)$ decreases the most from $f(\bm x_i)$ if $\Delta \bm x_i$ is chosen in the direction of $\bm M (\bm x_l-\bm x_j)$: $f(\bm x_i + \Delta \bm x_i)-f(\bm x_i)=2\Delta \bm x_i^T \bm M (\bm x_j-\bm x_l) = -2r \|\bm M (\bm x_l-\bm x_j)\|_2$. Therefore, in order to correctly classify the perturbed instance $\bm x_i+\Delta \bm x_i$, it is required that $f(\bm x_i + \Delta \bm x_i)$ is positive, that is, $\|\bm M (\bm x_l-\bm x_j)\|_2$ should be small. One way to reduce this value is by regularizing the spectral norm of $\bm M$. However, it is demanding for the metric to satisfy the large distance margin for all triplets and meanwhile keep a small spectral norm.

To achieve robustness against instance perturbation, we suggest an alternative way by maximizing the adversarial margin, defined as the distance between the training instance and its closest \emph{adversarial example}~\cite{szegedy2013intriguing}. More concretely, an adversarial example is a perturbed point whose nearest neighbor, identified based on the learned Mahalanobis distance, changes from an instance of the same class to one of a different class; consequently, it will be misclassified by the NN classifier and increase the risk of misclassification by $k$NN. In terms of previous notations, an adversarial example is a perturbed point $\bm x_i+\Delta \bm x_i$ such that $f(\bm x_i+\Delta \bm x_i)<0$. If all adversarial examples of an instance are far away from the instance itself, i.e., there is no $\Delta \bm x_i$ such that $\|\Delta \bm x_i\|_2 \leq r$ and $f(\bm x_i+\Delta \bm x_i)<0$, a high degree of robustness is achieved. Building on this rationale, we will first find the closest adversarial example and then push this point away from the training instance. Moreover, since the test instance can be regarded as a perturbed copy of training instances~\cite{xu2012robustness}, improving robustness on correctly classified training instances also helps enhance the generalization ability of the learned metric.

\subsection{Derivation of adversarial margin}
\label{subsec:support_point}

We start by deriving a closed-form solution to the closest adversarial example. Given a training instance $\bm x_i$ and the associated triplet constraint $(\bm x_i, \bm x_j, \bm x_l)$, we aim to find the closest point $\bm x_{i,{\rm min}}$ to $\bm x_i$ in the input space that lies on the decision boundary formed by $\bm x_j$ and $\bm x_l$ in the feature space. Note that closeness is defined in the input space and will be calculated using the Euclidean distance since we target at changes on the original feature of an instance; and that the decision boundary is found in the feature space since $k$NNs are identified by using the Mahalanobis distance. Mathematically, we can formulate the closest adversarial example $\bm x_{i,{\rm min}}$ as follows:
\begin{equation}
\begin{split}
    &\bm x_{i,{\rm min}}=\underset{\bm x'_i\in \mathbb{R}^p}{\arg\min} (\bm x'_i - \bm x_i)^T (\bm x'_i - \bm x_i) \quad \\ 
    &\text{s.t. } (\bm L \bm x'_i - \frac{\bm L \bm x_j + \bm L \bm x_l}{2})^T (\bm L \bm x_l - \bm L \bm x_j) = 0 
\end{split}
\label{eq:x_min_opt}
\end{equation}
The objective function of Eq.~\ref{eq:x_min_opt} corresponds to minimizing the Euclidean distance from the training instance $\bm x_i$. The constraint represents the decision boundary, which is the perpendicular bisector of points $\bm L \bm x_j$ and $\bm L \bm x_l$. In other words, it is a hyperplane that is perpendicular to the line joining points $\bm L \bm x_j$ and $\bm L \bm x_l$ and passes their midpoint $\frac{\bm L \bm x_j + \bm L \bm x_l}{2}$; all points on the hyperplane are equidistant from $\bm L \bm x_j$ and $\bm L \bm x_l$. 

Since Eq.~\ref{eq:x_min_opt} minimizes a convex quadratic function with an equality constraint, we can find an explicit formula for $\bm x_{i,{\rm min}}$ by using the method of Lagrangian multipliers; detailed derivation is provided in Section~\ref{sec:derivation} in the Supplementary Material: 
\begin{equation}
    \bm x_{i,{\rm min}} = \bm x_i +\frac{\left(\frac{\bm x_j + \bm x_l}{2} - \bm x_i\right)^T \bm M (\bm x_l - \bm x_j)}{(\bm x_l - \bm x_j)^T \bm M^2 (\bm x_l - \bm x_j)} \bm M (\bm x_l - \bm x_j). 
\label{eq:x_min_def}
\end{equation}

With the solution of $\bm x_{i,{\rm min}}$, we can now calculate the squared Euclidean distance between $\bm x_i$ and $\bm x_{i,{\rm min}}$:
\begin{equation}
    d^2_{\bm E}(\bm x_i, \bm x_{i,{\rm min}}) = \frac{\left( d^2_{\bm M}(\bm x_i, \bm x_l) - d^2_{\bm M}(\bm x_i, \bm x_j) \right)^2}{4 d^2_{\bm M^2}(\bm x_j, \bm x_l)}.
    \label{eq:adv_margin}
\end{equation}
For clarity, we will call $d_{\bm E}(\bm x_i, \bm x_{i,{\rm min}})$ the \emph{adversarial margin}, in contrast to the distance margin as in LMNN. It represents the maximum amount of tolerance for perturbation while retaining prediction correctness. The numerator of Eq.~\ref{eq:adv_margin} is the square of the standard distance margin, and the denominator is the squared $L_2$-norm of $\bm M (\bm x_l-\bm x_j)$. Therefore, in order to achieve a large adversarial margin, the metric should push $\bm x_l$ away from the neighborhood of $\bm x_i$ by expanding the distance in the direction that has a small correlation with $\bm x_l - \bm x_j$ (the optimal direction is orthogonal to $\bm x_l - \bm x_j$). 

\paragraph*{Remark 1} The objective function in Eq.~\ref{eq:x_min_opt} defines a hypersphere in the input space, which characterizes perturbations of equal magnitude in all directions, e.g., isotropic Gaussian noise. To model heterogeneous and correlated perturbation, we can extend the objective function by defining an arbitrary oriented hyperellipsoid, as discussed in Section~\ref{sec:derivation} in the Supplementary Material.

\subsection{Metric learning via minimizing the perturbation loss} 
\label{subsec:optimization}

To improve robustness of distance metric, we design a perturbation loss to promote an increase in the adversarial margin. Two situations need to be distinguished here. Firstly, when the nearest neighbor of $\bm x_i$ is an instance from the same class, we will penalize a small adversarial margin by using the hinge loss $[\tau^2 - d_{\bm E}^2(\bm x_i, \bm x_{i,{\rm min}})]_+$. The reasons are that (a) the adversarial margin is generally smaller for hard instances that are close to the class boundary in contrast to those locating far away and (b) it is these hard instances that are more vulnerable to perturbation and demand an improvement in their robustness. Therefore, we introduce $\tau$ for directing attention to hard instances and controlling the desired margin. Secondly, in the other situation where the nearest neighbor of $\bm x_i$ belongs to a different class, metric learning should focus on satisfying the distance requirement specified in the triplet constraint. In this case, we simply assign a large penalty of $\tau^2$ to promote a non-increasing loss function. Integrating these two situations, we propose the following perturbation loss: 
\begin{equation}
\begin{split}
    J_\text{P} =& \frac{1}{|\mathcal{R}|}\sum_{\mathcal{R}}
    \big \{ 
    [\tau^2 - \tilde{d}_{\bm E}^2(\bm x_i, \bm x_{i,{\rm min}})]_+ \mathds{1}[d^2_{\bm M}(\bm x_i, \bm x_l) > d^2_{\bm M}(\bm x_i, \bm x_j)] \\
    +&
    \tau^2 \mathds{1}[d^2_{\bm M}(\bm x_i, \bm x_l) \leq d^2_{\bm M}(\bm x_i, \bm x_j)]
    \big\} ,
\end{split}
\label{eq:perturbation_loss}
\end{equation}
where $\sum_\mathcal{R}$ is an abbreviation for $\sum_{(\bm x_i, \bm x_j, \bm x_l) \in \mathcal{R}}$. To prevent the denominator of Eq.~\ref{eq:adv_margin} from being zero, which may happen when different-class instances $\bm x_j$ and $\bm x_l$ are close to each other, we add a small constant $\epsilon$ ($\epsilon$=1e-10) to the denominator; that is, $\tilde{d}^2_{\bm E}(\bm x_i, \bm x_{i,{\rm min}}) = \frac{\left( d^2_{\bm M}(\bm x_i, \bm x_l) - d^2_{\bm M}(\bm x_i, \bm x_j) \right)^2}{4 \left( d^2_{\bm M^2}(\bm x_j, \bm x_l)+\epsilon\right)}$.

The proposed perturbation loss can be readily included in the objective function of any metric learning methods and is particularly useful to triplet-based methods. When the same triplet set is used for supervising metric learning and deriving adversarial examples, our method can encourage the triplets to meet the distance margin by learning a discriminative metric. For this reason, we adapt LMNN as an example for its wide use and effective classification performance. The objective function of LMNN with the perturbation loss is as follows:
\begin{equation}
\begin{split}
    \min_{\bm M \in \mathbb{S}^p_+} J &= J_\text{LMNN} + \lambda J_\text{P},\\
    J_\text{LMNN} &= (1-\mu) \frac{1}{|\mathcal{S}|} \sum_\mathcal{S} d^2_{\bm M}(\bm x_i, \bm x_j) \\
    &+ \mu \frac{1}{|\mathcal{R}|} \sum_{\mathcal{R}} \left[1+ d^2_{\bm M}(\bm x_i, \bm x_j) - d^2_{\bm M}(\bm x_i, \bm x_l) \right]_+ ,
\end{split}
\label{eq:obj}
\end{equation}
where $\sum_\mathcal{S}$ stands for $\sum_{(\bm x_i, \bm x_j) \in \mathcal{S}}$. The weight parameter $\lambda>0$ controls the importance of perturbation loss ($J_\text{P}$) relative to the loss function of LMNN ($J_\text{LMNN}$). $\mu \in (0,1)$ balances the impacts between pulling together target neighbors and pushing away impostors. 

We adopt the projected gradient descent algorithm to solve the above optimization problem. The gradient of $J_\text{P}$ and $J_\text{LMNN}$ are given as follows: 
\begin{equation}
\begin{split}
    &\frac{\partial J_\text{P}}{\partial \bm M} 
    = \frac{1}{|\mathcal{R}|} \sum_\mathcal{R} \alpha_{ijl} \bigg\{ \frac{d^2_{\bm M}(\bm x_i, \bm x_l) - d^2_{\bm M}(\bm x_i, \bm x_j)}{2\left(d^2_{\bm M^2} (\bm x_j, \bm x_l)+\epsilon\right)} (\bm X_{ij} - \bm X_{il})\\
    &\hspace{2.3em}+ \frac{\left(d^2_{\bm M}(\bm x_i, \bm x_l) - d^2_{\bm M}(\bm x_i, \bm x_j)\right)^2}{4\left(d^2_{\bm M^2} (\bm x_j, \bm x_l)+\epsilon\right)^2}(\bm M \bm X_{jl} + \bm X_{jl} \bm M) \bigg\},\\
    &\frac{\partial J_\text{LMNN}}{\partial \bm M} = \frac{1-\mu}{|\mathcal{S}|}\sum_\mathcal{S} \bm X_{ij} + \frac{\mu}{|\mathcal{R}|} \sum_\mathcal{R} \beta_{ijl} (\bm X_{ij} - \bm X_{il}),
\end{split}
\label{eq:grad}
\end{equation}
where $\alpha_{ijl} = \mathds{1}[d^2_{\bm M}(\bm x_i, \bm x_l) > d^2_{\bm M}(\bm x_i, \bm x_j), \tilde{d}_{\bm E}(\bm x_i, \bm x_{i,{\rm min}}) \leq \tau]$, $\allowdisplaybreaks \beta_{ijl} = \mathds{1}[1+ d^2_{\bm M}(\bm x_i, \bm x_j) - d^2_{\bm M}(\bm x_i, \bm x_l) \geq 0]$; $\bm X_{ij} = (\bm x_i - \bm x_j) (\bm x_i - \bm x_j)^T$ and $\bm X_{il}, \bm X_{jl}$ are defined similarly. The gradient of $J_\text{P}$ is a sum of two descent directions. The first direction $\bm X_{ij}-\bm X_{il}$ agrees with LMNN, indicating that our method updates the metric toward better discrimination in a weighted manner. The second direction $\bm M \bm X_{jl} + \bm X_{jl} \bm M$ controls the scale of $\bm M$; the metric will descend at a faster pace in the direction of a larger correlation between $\bm M$ and $\bm X_{jl}$. This suggests our method functions as a data-dependent regularization. Let $\bm M^t$ denote the Mahalanobis matrix learned at the $t$th iteration. The distance matrix will be updated as
\begin{equation*}
    \bm M^{t+1} = \bm M^t - \gamma \left( \frac{\partial J_\text{LMNN}}{\partial \bm M^t} + \lambda \frac{\partial J_\text{P}}{\partial \bm M^t} \right),
\end{equation*}
where $\gamma$ denotes the learning rate. Following~\cite{weinberger2009distance}'s work, $\gamma$ is increased by $1\%$ if the loss function decreases and decreased by $50\%$ otherwise. To guarantee the PSD property, we factorize $\bm M^{t+1}$ as $\bm V \bm \Lambda \bm V^T$ via eigendecomposition and truncate all negative eigenvalues to zero, i.e. $\bm M^{t+1} = \bm V \max(\bm \Lambda,0) \bm V^T$.

\paragraph*{Remark 2} The proposed perturbation loss is a generic approach to improving robustness against possible perturbation. In Section~\ref{sec:CR_extension} in the Supplementary Material, we illustrate examples of incorporating the perturbation loss into two different types of triplet-based methods, sparse compositional metric learning (SCML)~\cite{shi2014sparse} and proxy neighborhood component analysis (ProxyNCA++)~\cite{teh2020proxynca++}. SCML revises the structure of the Mahalanobis distance by representing it as a sparse and non-negative combination of rank-one basis elements, which typically results in less number of parameters to be estimated. ProxyNCA++ revises the construction of triplet constraints by replacing nearest instances $\bm x_j$ and $\bm x_l$ with nearest proxy points. The proxies are learned to represent each class, and the resulting method is shown to generalize well on small datasets~\cite{snell2017prototypical}, robust to outliers and noisy labels~\cite{kim2020proxy}, and improves computational efficiency on large-scale datasets. 

\paragraph*{Remark 3}  Learning a distance metric for extremely high-dimensional data will result in a large number of parameters to be estimated and potentially suffer from overfitting. In order to reduce the input dimensionality, PCA is often applied to pre-process the data prior to metric learning~\cite{weinberger2009distance,huo2016robust}. In Section~\ref{sec:derivation_HD} in the Supplementary Material, we extend the proposed method such that the distance metric learned in the low-dimensional PCA subspace could still achieve robustness against perturbation in the original high-dimensional input space. The decision boundary of NN classifier (i.e., the constraint of Eq.~\ref{eq:x_min_opt}) is revised in order to take account of the linear transformation matrix induced by the Mahalanobis distance and that of PCA. The proposed extension will be evaluated in Section~\ref{subsec:exp_HD}.

\subsection{Generalization benefit}
\label{subsec:bound}

From the perspective of algorithmic robustness~\cite{xu2012robustness}, enlarging the adversarial margin could potentially improve the generalization ability of triplet-based metric learning methods. The following generalization bound, i.e., the gap between the generalization error $\mathcal{L}$ and the empirical error $\ell_{emp}$, follows from the pseudo-robust theorem of~\cite{bellet2015robustness}. Preliminaries and derivations are given in Section~\ref{sec(app):generalization} in the Supplementary Material.
\begin{thm}
\label{thm:bound}
    Let $\bm M^\ast$ be the optimal solution to Eq.~\ref{eq:obj}. Then for any $\delta>0$, with probability at least $1-\delta$ we have:
    \begin{equation}
    \begin{split}
        &|\mathcal{L}(\bm M^\ast)-\ell_{emp}(\bm M^\ast)| \\
        \leq& \frac{\hat{n}(t_{\bm s})}{n^3}+ B\bigg(\frac{n^3 - \hat{n}(t_{\bm s})}{n^3} + 3\sqrt{\frac{2K \ln 2+2\ln 1/\delta}{n}}\bigg),
    \end{split}
    \label{eq:bound}
    \end{equation}
    where $\hat{n}(t_{\bm s})$ denotes the number of triplets whose adversarial margins are larger than $\tau$, $B$ is a constant denoting the upper bound of the loss function (i.e., Eq.~\ref{eq:obj}), and $K$ denotes the number of disjoint sets that partition the input-label space and equals to $|\mathcal{Y}|(1+\frac{2}{\tau})^p$.
\end{thm}

Enlarging the desired adversarial margin $\tau$ will affect two quantities in Eq.~\ref{eq:bound}, namely $K$ and $\hat{n}(t_{\bm s})$. First, since $K$ equals to $|\mathcal{Y}|(1+\frac{2}{\tau})^p$, increasing $\tau$ will cause $K$ to decrease at a polynomial rate of the input dimensionality $p$. Moreover, as the right-hand side of Eq.~\ref{eq:bound} is a function of $K$ ($\mathcal{O}(K^{1/2})$), this means that the upper bound of generalization gap reduces at a rate of $p^{1/2}$. Hence, for datasets with a relative large number of features, a small improvement in the adversarial margin can greatly benefit the generalization ability of the learned metric. 

Secondly, when $\tau$ increases, less triplets will satisfy the condition that their adversarial margin is larger than $\tau$; that is, $\hat{n}(t_{\bm s})$ decreases with $\tau$. Meanwhile, since $B>1$, the upper bound is a decreasing function of $\hat{n}(t_{\bm s})$. Therefore, enlarging $\tau$ leads to an increase in the upper bound. However, the rate of such increase depends on the datasets. For example, if most instances in the dataset are well separated and have a margin in the original input space, enlarging the desired adversarial margin $\tau$ will not have a large impact on $\hat{n}(t_{\bm s})$, the upper bound, and thus the generalization gap. 

In summary, for datasets with many features and most instances being separable, we expect an improvement in the generalization ability of the learned distance metric from enlarging the adversarial margin.

\section{Related Work}
\label{sec:related_work}

\subsection{Robust metric learning}
To make machine learning models more secure and trustworthy, robustness to input perturbations is a crucial dimension~\cite{floridi2019establishing}. More importantly, designing such robust metric learning algorithms is particularly vital to safety-critical applications, such as healthcare~\cite{suo2018multi}, network intrusion detection~\cite{aliakbarisani2019data}, and surveillance systems based on faces~\cite{guillaumin2009you}, gaits~\cite{ben2016distance} and other biometric traits~\cite{omara2018metric}.

Existing approaches to improving the robustness of Mahalanobis distances can be categorized into four main types. The first type of method imposes structural assumption or regularization over $\bm M$ so as to avoid overfitting~\cite{jin2009regularized,lim2013robust,law2014fantope,huo2016robust,luo2018matrix,liu2019learning}. Methods with structural assumption are proposed for classifying images and achieve robustness by exploiting the structural information of images; however, such information is generally unavailable in the symbolic datasets that will be studied in this paper. Regularization-based methods are proposed to reduce the risk of overfitting to feature noise in the training set. Our proposal, which is aimed to withstand test-time perturbation, does not conflict with these methods and can be combined with them to learn a more effective and robust distance metric; an example is shown in Section~\ref{subsec:exp_HD}. 

The second type of method adopts loss functions that are less sensitive to outlier samples or noisy labels. In most metric learning methods, loss functions are founded on the squared $L_2$-norm distance for computational efficiency. However, such choice may be sensitive to outliers. To overcome this limitation, several remedies have been proposed, such as using $L_1$-norm distances~\cite{wang2014robust} and metric based on the signal-to-noise ratio~\cite{yuan2019signal}, or replacing the square function with the maximum correntropy criterion~\cite{xu2018new}. 

The third type of method studies robustness to training noise~\cite{ye2017learning,qian2018large}. These methods explicitly model the noise distribution or identify clean latent examples, and consequently, use the expected Mahalanobis distance to adjust the value of the distance margin for each triplet. Our method can also be viewed as imposing a data-dependent and dynamic margin -- to achieve the same adversarial margin, triplets that have a higher correlation between $\bm x_l-\bm x_j$ and the metric $\bm M$ should satisfy a larger distance margin. However, the focus of our work is orthogonal to the aforementioned two types of method.

The last type of method generates hard instances through adversarial learning and trains a metric to fare well in the new hard problem~\cite{chen2018adversarial,duan2018deep}. While sharing the aim of improving metric robustness, our method is intrinsically different from them. Their methods approach the task at a data-level, where real examples are synthesized based on the criterion of incurring large losses. Our method tackles perturbation at a model-level, where a loss function is derived by considering the definition of robustness with respect to the decision maker $k$NN. By preventing change in the nearest neighbor in a strict manner, our method is capable of obtaining a certification on adversarial margin. 

\subsection{Adversarial robustness of deep metric learning}

More recently, deep metric learning has been investigated intensively, which replaces the linear projection induced by the Mahalanobis distance with deep neural networks. While deep neural networks improve the discriminability between classes, they are found to be non-robust and vulnerable to adversarial examples. Robust optimization~\cite{ben2009robust,szegedy2013intriguing} is one of the most effective approaches to improving adversarial robustness, which trains the network to be robust against adversarial perturbations that are mostly constructed via gradient-based optimization; \cite{panum2021adversarial} adapts it to deep metric learning by considering the interdependence between data points in pairwise or triplet constraints. Another way to enhance robustness and generalization ability is by attaining a large margin in the input space, which dates back to support vector machines~\cite{cortes1995support} and inspires this work. Due to the hierarchical nonlinear nature of deep networks, the input-space margin cannot be computed exactly and a variety of approximations have been proposed~\cite{an2015contractive,elsayed2018large,yan2019adversarial,ding2020mma}. In this work, we investigate such margin in the framework of metric learning, defines it specifically with respect to the $k$NN classifier, and provide an exact and analytical solution to the margin. The analytical solution to the margin provides fascinating insights into essential factors for the robustness of distance metrics.

\subsection{Adversarial robustness of $k$NN classifiers}
While the notion of adversarial examples applies to $k$NN classifiers, existing methods for deep neural networks cannot be implemented directly due to the non-differential nature of the classifier. \cite{papernot2016transferability} and \cite{sitawarin2019robustness} propose continuous substitutes of $k$NN, from which gradient-based adversarial examples can be constructed to attack the classifier. \cite{wang2019evaluating} formulates a series of quadratic programming (QP) problems and proposes an efficient algorithm to search exhaustively over all training samples and compute the minimal adversarial perturbation for the 1-NN classifier. In addition, the dual solution to these QP problems can be used for robustness verification. \cite{yang2020robustness} proposes to improve adversarial robustness for $k$NN by pruning the training set in order to satisfy the condition defined through the robustness radius, i.e., the norm of the minimal adversarial perturbation. Our work also aims to robustify $k$NN, but achieves it through enlarging the adversarial margin. 

\section{Experiments}
\label{sec:exp}

In this section, we first present two toy examples to illustrate the difference in the learning mechanisms of LMNN and the proposed method dubbed LMNN-PL. Next, we compare LMNN-PL with state-of-the-art methods on 16 benchmark datasets (13 low/medium-dimensional and three high-dimensional), and investigate the relationship between adversarial margin, generalization ability and robustness. Finally, the computational aspect of our method is discussed. 

\subsection{Comparisons between LMNN and LMNN-PL}
\label{subsec:LMNN_vs_CR}

\begin{figure*}[t]
\centering
\includegraphics[width=.45\linewidth]{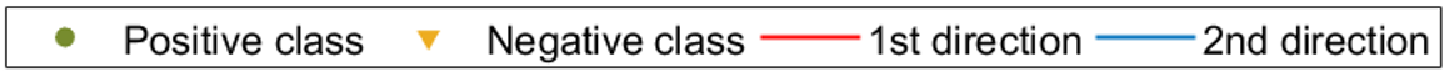}\\\vspace{.5em}
\subfloat[][\hangindent=1.9em\tiny{$\bm M = \begin{bmatrix}1 & 0 \\ 0 & 1 \end{bmatrix} \hspace{3em}$ $\bar{d}_{\bm E}(\bm x_i, \bm x_{i,{\rm min}})$ $=0.17$}]{{\includegraphics[width=.325\linewidth]{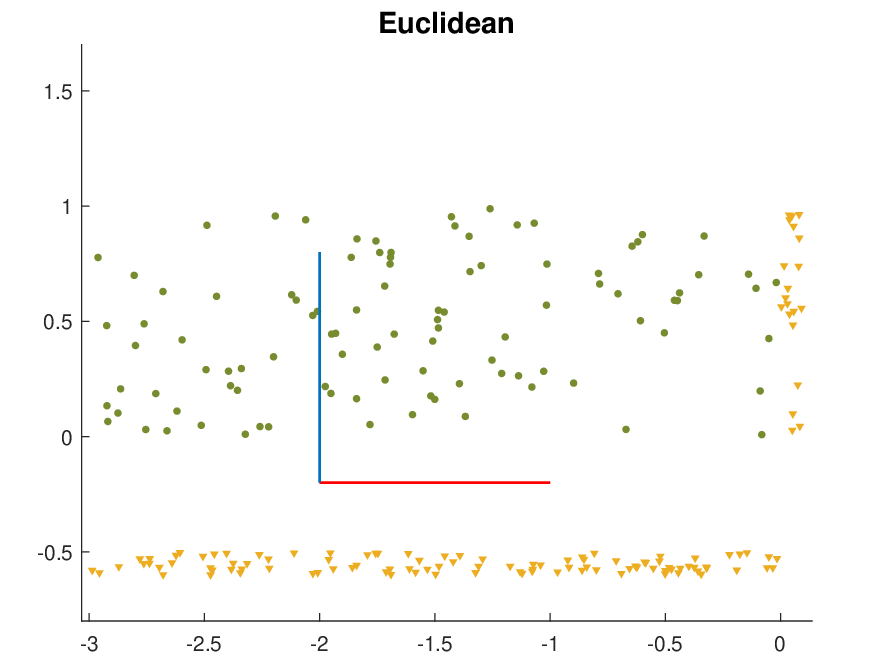}}
\label{fig:s1_Euc}}
\hfil
\subfloat[][\hangindent=1.9em\tiny{$\bm M = \begin{bmatrix} 10.2 & 3.5 \\ 3.5 & 7.8 \end{bmatrix} \hspace{1em}$ $\bar{d}_{\bm E}(\bm x_i, \bm x_{i,{\rm min}})=0.15$}]{{\includegraphics[width=.325\linewidth]{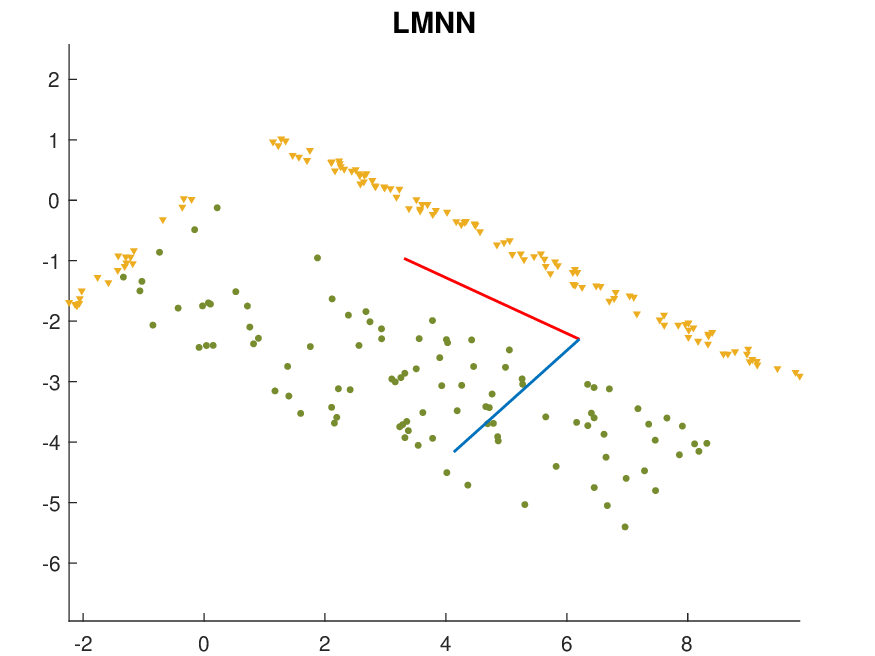}}
\label{fig:s1_LMNN}}
\hfil
\subfloat[][\hangindent=1.9em \tiny{$\bm M = \begin{bmatrix} 2.1 & 0.6 \\ 0.6 & 3.1 \end{bmatrix} \hspace{1em}$ $\bar{d}_{\bm E}(\bm x_i, \bm x_{i,{\rm min}})=0.16$}]{{\includegraphics[width=.325\linewidth]{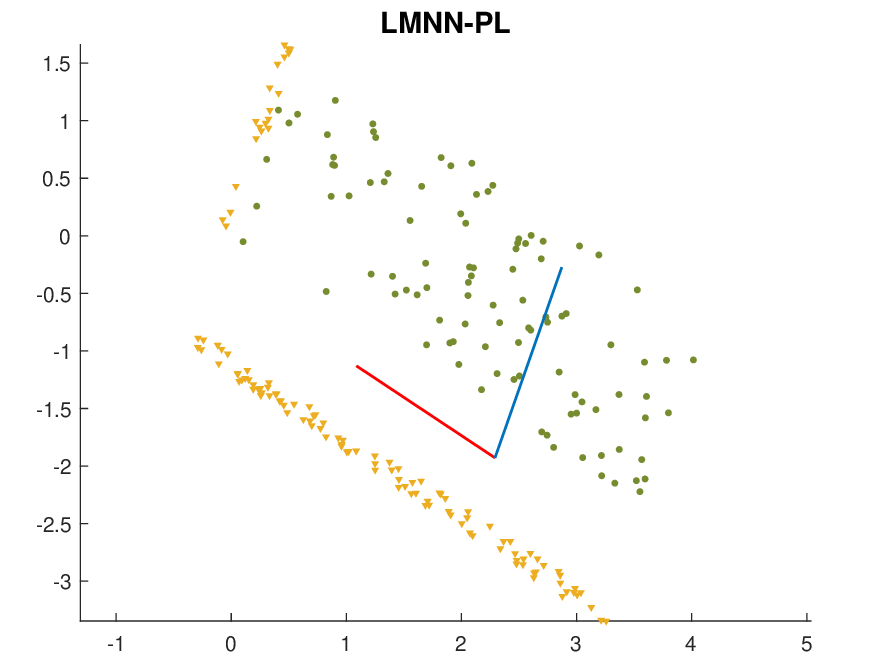}}
\label{fig:s1_LMNNCR}}
\caption{Comparison of learning mechanisms of LMNN and LMNN-PL when features exhibit different separability.}
\label{fig:simulation1}
\vspace{1em}
\end{figure*}

We design two experiments to compare the metrics learned with the objective of enhancing class discriminability and of certified robustness. In the first example, a two-dimensional binary classification dataset is simulated, as shown in Fig.~\ref{fig:s1_Euc}. The positive class includes 100 instances drawn uniformly from $[-3,0]$ in the horizontal (abbr. 1st) direction  and $[0,1]$ in the vertical (abbr. 2nd) direction. The negative class consists of two clusters, where the first cluster includes 100 instances drawn from $U(-3,0)$ and $U(-0.6,-0.5)$ in the 1st and 2nd directions respectively, and the second cluster includes 20 instances drawn from $U(0,0.1)$ and $U(0,1)$ in the two directions respectively. By design, instances of positive and negative classes can be separated in both directions, while the separability in the 1st direction is much smaller than the 2nd direction. Figs.~\ref{fig:s1_LMNN} and~\ref{fig:s1_LMNNCR} show the instances in the projected feature space with metrics learned from LMNN and LMNN-PL, respectively; the projection direction is indicated by the unit vector of red and blue lines; and the metric and the average of adversarial margins ($\bar{d}_{\bm E}(\bm x_i, \bm x_{i,{\rm min}})$) are given in the caption. The objective of LMNN is to satisfy the distance margin. Thus, it expands the distance in both directions. Moreover, since the 1st direction has a small separability in the original instance space, this direction is assigned with a larger weight. In contrast, LMNN-PL controls the scale of $\bm M$. Moreover, a notable difference is that the 2nd direction is assigned with a larger weight than the 1st direction, which is again caused by the small separability in the 1st direction. As any perturbation in the 1st direction is highly likely to result in a misclassification, the proposed method diverts more attention to robust features, i.e., the 2nd direction. Due to the easiness of the task, all metrics lead to the same classification accuracy of 99.09\% on a separate test set. 

\begin{figure*}[t]
\centering
\includegraphics[width=.53\linewidth]{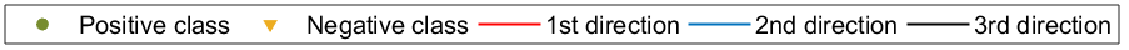}\\\vspace{.5em}
\subfloat[][\hangindent=1.9em\tiny{$\setlength\arraycolsep{2pt} \bm M = \begin{bmatrix}
    1 & 0 & 0 \\ 0 & 1 & 0\\
    0 & 0 & 1 \end{bmatrix} \hspace{3em}$ $\bar{d}_{\bm E}(\bm x_i, \bm x_{i,{\rm min}})$=0.83}]{{\includegraphics[width=.325\linewidth]{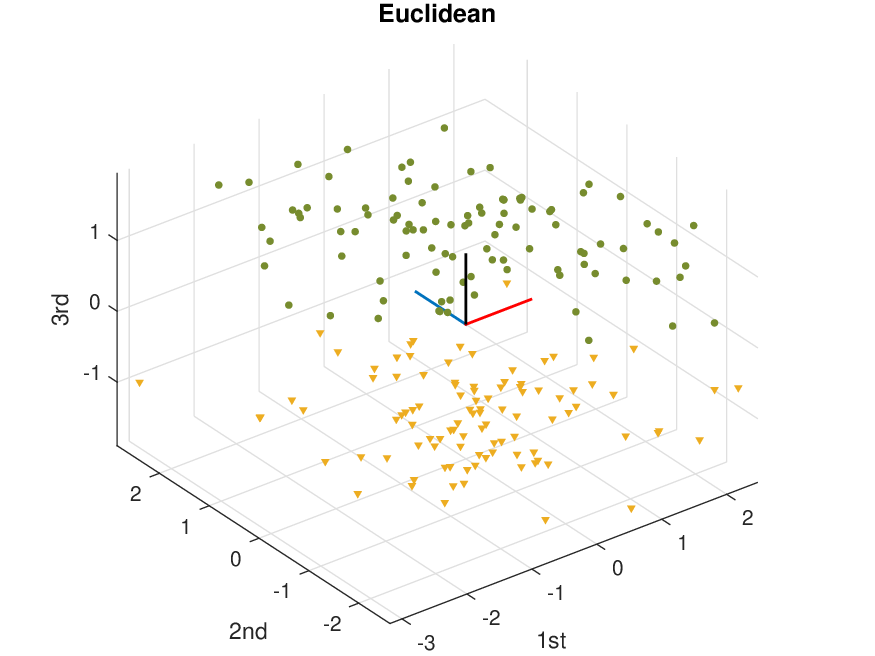}}
\label{fig:s2_Euc}}
\hfil
\subfloat[][\hangindent=1.9em\tiny{ $\setlength\arraycolsep{2pt} \bm M = \begin{bmatrix} 
    22 & 23 & -17 \\
    23 & 28 & -19\\
    -17 & -19 & 14
    \end{bmatrix} \hspace{1em}$ $\bar{d}_{\bm E}(\bm x_i, \bm x_{i,{\rm min}})$=0.05}]{{\includegraphics[width=.325\linewidth]{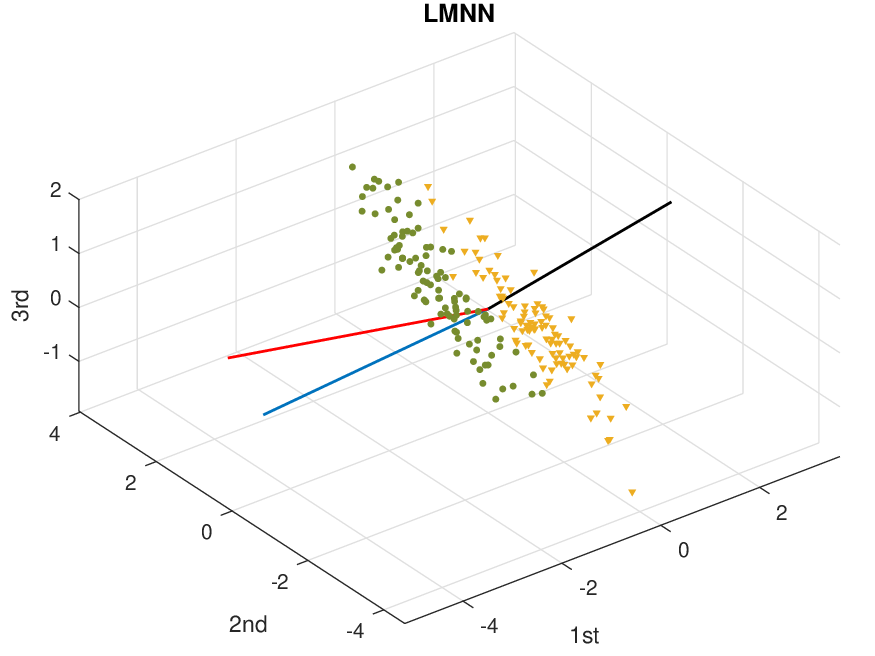}}
\label{fig:s2_LMNN}}
\hfil
\subfloat[][\hangindent=1.9em\tiny{$\setlength\arraycolsep{1pt} \bm M = \begin{bmatrix}
    0.45 & 0.45 & -0.08 \\ 
    0.45 & 0.45 & -0.06 \\
    -0.08 & -0.06 & 0.85 \end{bmatrix}$ $\bar{d}_{\bm E}(\bm x_i, \bm x_{i,{\rm min}})$=0.70}]{{\includegraphics[width=.325\linewidth]{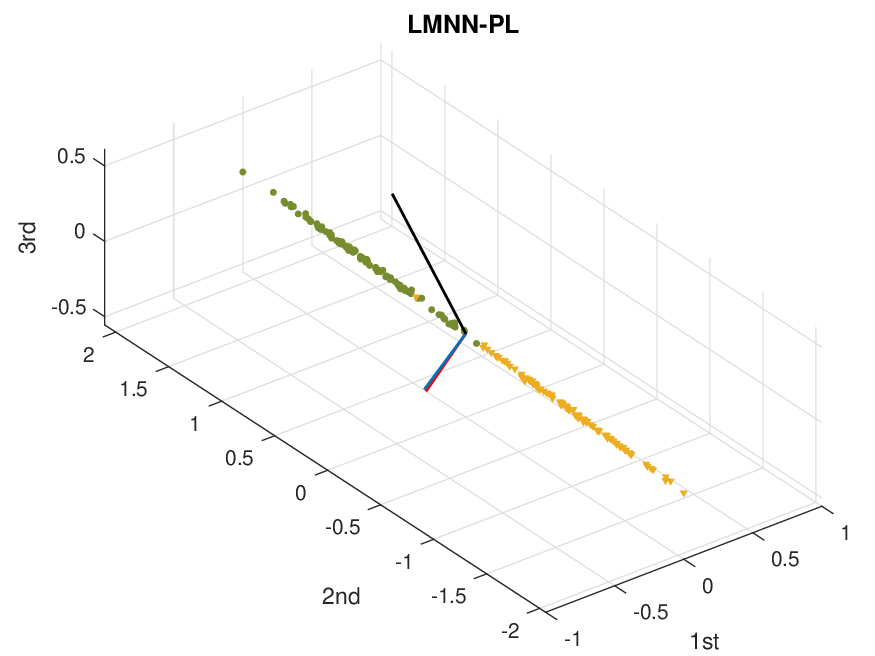}}
\label{fig:s2_LMNNCR}}
\caption{Comparison of learning mechanisms of LMNN and LMNN-PL when confronting the problem of multicollinearity.}
\label{fig:simulation2}
\end{figure*}

In the second example, we simulate a three-dimensional binary classification dataset, as shown in Fig.~\ref{fig:s2_Euc}. Each class includes 100 instances. The first two dimensions are drawn from multivariate Gaussian distributions with $\bm \mu_p=[0.45,0.45]$, $\bm \mu_n=[-0.45,-0.45]$, $\bm \Sigma_p=\bm \Sigma_n=\big[\begin{smallmatrix} 1 & -0.9\\ -0.9 & 1  \end{smallmatrix}\big]$; the third dimension equals the sum of the first two dimensions, plus white Gaussian noise with standard deviation of 0.01. By design, the dataset exhibits the problem of strong multicollinearity. This issue has little influence on LMNN as the data is nearly separable in all directions. However, it will affect the adversarial margin. Specifically, if the metric assigns equal weights to all dimensions, then the perturbation should be small in all directions so as to guarantee that the perturbed instance stays on the correct side of the decision boundary. In contrast, if the metric assigns weights only to the third dimension, then the perturbation in the first two dimensions will not cause any change in the learned feature space and hence a larger magnitude of perturbation can be tolerated. This expectation is supported by the empirical result in Fig.~\ref{fig:s2_LMNNCR}, where the distance in the third dimension is more important than the first two dimensions. LMNN achieves an accuracy of 95.50\% and our method achieves an accuracy of 96.00\%.

In summary, our method learns a discriminative metric, and meanwhile, imposes a data-dependent regularization on the metric. It also achieves larger adversarial margins than LMNN, demonstrating the effectiveness of the proposed perturbation loss. 

\subsection{Experiments on UCI data}

\subsubsection{Data description and experimental setting}

In this experiment, we study 13 datasets from UCI machine learning repository~\cite{Dua:2019}. Information on sample size, feature dimension and class information is listed in Table~\ref{tab:data} in the Supplementary Material. All datasets are pre-processed with mean-centering and standardization, followed by $L_2$ normalization to unit length. To evaluate the performance, we use 70-30\% training-test partitions and report the average result over 20 rounds of random split. The only exception is the Credit dataset, where we only run the experiment once as the sample size is relatively large. 

We evaluate the effectiveness of the proposed perturbation loss by incorporating it into LMNN, SCML, and ProxyNCA++ (abbreviated to PNCA); the resulting methods are denoted by LMNN-PL, SCML-PL and PNCA-PL, respectively. In addition, we conduct a thorough study by setting LMNN as the backbone and comparing LMNN-PL with two types of methods. First, we consider different regularizers on $\bm M$. Specifically, we replace the regularizer in LMNN from $\sum_\mathcal{S} d^2_{\bm M}(\bm x_i, \bm x_j)$ to the log-determinant divergence~(LDD)~\cite{davis2007information}, which encourages learning a metric toward the identity matrix, the capped trace norm~(CAP)~\cite{huo2016robust}, which encourages a low-rank matrix, and the spectral norm~(SN), which has been used to improve adversarial robustness of deep neural networks~\cite{bartlett2017spectrally}. Second, we compare with the robust metric learning method DRIFT~\cite{ye2017learning}, which models the perturbation distribution explicitly.

Hyperparameters of LMNN-PL are tuned via random search~\cite{bergstra2012random}. We randomly sample 50 sets of values from the following ranges: $\mu \in U(0.1,0.9)$, $\tau \in U\left(0,P_{90\%}\{d_{\bm E}(\bm x_i, \bm x_{i,{\rm min}})\}\right)$, $\lambda \in U(0, 4/\tau^2)$. $U(a,b)$ denotes the uniform distribution. $P_{k\%}\{d_{\bm E}(\bm x_i, \bm x_{i,{\rm min}})\}$ denotes the $k$th percentile of $d_{\bm E}(\bm x_i, \bm x_{i,{\rm min}})$, where the distance is calculated for all $i$ in the triplet constraints with respect to the Euclidean distance. Setting the upper bound of the desired margin $\tau$ via the percentile avoids unnecessary large values, matching our intention to enlarge the adversarial margin primarily for hard instances. The upper bound of the weight parameter $\lambda$ depends on the realization of $\tau$ to ensure that magnitudes of perturbation loss and LMNN loss are at the same level. The optimal hyperparameters from five-fold cross-validation on the training data or a separate validation set are used to learn the metric. SCML-PL and PNCA-PL are tuned in a similar manner. More details on the training procedure of the proposed and other methods are given in Section~\ref{subsec(app):exp_setting} in the Supplementary Material. The MATLAB code for our method is available at
\url{http://github.com/xyang6/LMNNPL.}

For LMNN-based and SCML-based methods, we use 3NN as the classifier; for PNCA-based methods, we use the nearest prototype classifier. Classification accuracy is used as the evaluation criterion, except for two highly imbalanced datasets (Ecoli and Yeast), G-means is used.  

\subsubsection{Evaluation on classification performance}

Table~\ref{tab:clean} reports the mean value and standard deviation of classification accuracy or G-means for imbalanced datasets (indicated by an asterisk). LMNN-PL outperforms LMNN on 12 out of 13 datasets. Among the methods with LMNN as the backbone, our method achieves the highest accuracy on 8 datasets and second highest accuracy on the 4 datasets. These experimental results demonstrate the benefit of perturbation loss to generalization of the learned metric. 
Similarly, we see that SCML-CL outperforms or performs equally well with SCML on 9 datasets. The advantage of PNCA-PL becomes less distinct as it is superior to PNCA only on 7 datasets. However, this is fairly reasonable as the decision boundary formed by very few proxies is much smoother than the one from 3NN and hence the method is less likely to overfit to training data. 

\begin{sidewaystable}
\centering
\small
\setlength\tabcolsep{2pt}
\caption{Classification accuracy (or G-means indicated by an asterisk next to the dataset name; mean$\pm$standard deviation) of 3NN with different metric learning methods on clean datasets.}
\label{tab:clean}
\begin{threeparttable}
\begin{tabular}{lccc cccc cccc ccc}
\toprule
&&& \multicolumn{6}{c}{LMNN-based}&& \multicolumn{2}{c}{SCML-based}&& \multicolumn{2}{c}{PNCA-based}\\
\cmidrule{4-9} \cmidrule{11-12} \cmidrule{14-15}
Dataset & Euclidean &  & LMNN & LDD & CAP & SN & DRIFT & LMNN-PL &  & SCML & SCML-PL &  & PNCA & PNCA-PL\\
\midrule
Australian & 82.76$\pm$2.38 &  & 83.70$\pm$2.43 & 84.18$\pm$2.37 & 83.97$\pm$2.45 & 83.77$\pm$2.50 & \textbf{84.47$\pm$2.02} & \textbf{84.47$\pm$1.63} &  & \textbf{84.76$\pm$2.08} & 84.42$\pm$2.18 &  & \textbf{86.30$\pm$1.82} & 86.15$\pm$2.24
\\
Breast cancer & 97.17$\pm$1.33 &  & \textbf{97.12$\pm$1.25} & 96.95$\pm$1.51 & 97.00$\pm$1.08 & \underline{97.05$\pm$1.32} & 96.98$\pm$1.16 & 97.02$\pm$1.30 &  & 97.00$\pm$1.09 & \textbf{97.07$\pm$1.24} &  & 97.02$\pm$1.30 & \textbf{97.05$\pm$1.24}
\\
Ecoli* & 85.86$\pm$8.10 &  & 86.42$\pm$7.94 & 85.29$\pm$9.78 & 83.54$\pm$10.09 & 86.44$\pm$7.95 & \underline{86.45$\pm$6.54} & \textbf{87.04$\pm$7.38} &  & 85.53$\pm$7.06 & \textbf{86.69$\pm$6.58} &  & 84.80$\pm$6.84 & \textbf{85.54$\pm$5.60}
\\
Fourclass & 75.12$\pm$2.35 &  & 75.10$\pm$2.31 & \textbf{75.15$\pm$2.32} & 75.02$\pm$2.48 & 75.10$\pm$2.31 & 75.08$\pm$2.34 & \underline{75.12$\pm$2.35} &  & 75.10$\pm$2.27 & \textbf{75.12$\pm$2.35} &  & \textbf{75.39$\pm$2.21} & 72.97$\pm$5.36
\\
Haberman & 72.25$\pm$4.41 &  & 72.19$\pm$3.89 & \underline{72.42$\pm$3.95} & 71.52$\pm$3.54 & 72.30$\pm$4.57 & 72.02$\pm$3.94 & \textbf{72.64$\pm$4.29} &  & \textbf{72.75$\pm$3.79} & 72.36$\pm$4.38 &  & 75.28$\pm$3.55 & \textbf{75.67$\pm$3.85}
\\
Iris & 87.11$\pm$4.92 &  & 87.11$\pm$5.08 & \textbf{87.67$\pm$4.70} & 86.67$\pm$5.49 & 87.22$\pm$5.24 & 85.89$\pm$4.46 & \underline{87.33$\pm$4.73} &  & 86.89$\pm$6.40 & \textbf{87.44$\pm$5.31} &  & \textbf{84.44$\pm$6.29} & 83.22$\pm$5.97
\\
Segment & 94.79$\pm$0.65 &  & 95.31$\pm$0.89 & 95.58$\pm$0.81 & 95.51$\pm$0.70 & 95.38$\pm$0.83 & \textbf{95.75$\pm$0.65} & \underline{95.64$\pm$0.83} &  & 92.61$\pm$6.65 & \textbf{93.95$\pm$1.47} &  & \textbf{94.73$\pm$0.93} & 94.52$\pm$0.99
\\
Sonar & 85.16$\pm$4.19 &  & 86.67$\pm$4.10 & \underline{87.22$\pm$3.90} & \underline{87.22$\pm$4.38} & 86.67$\pm$4.04 & 86.19$\pm$4.43 & \textbf{87.78$\pm$3.53} &  & 82.38$\pm$4.15 & \textbf{84.13$\pm$4.61} &  & 83.25$\pm$5.95 & \textbf{83.65$\pm$4.83}
\\
Voting & 93.78$\pm$1.76 &  & 95.80$\pm$1.78 & 95.80$\pm$1.41 & \underline{95.92$\pm$1.45} & 95.84$\pm$1.74 & 95.31$\pm$1.32 & \textbf{96.15$\pm$1.56} &  & 95.84$\pm$1.58 & \textbf{96.26$\pm$1.28} &  & \textbf{95.84$\pm$1.65} & 95.65$\pm$1.66
\\
WDBC & 96.29$\pm$1.61 &  & \underline{96.99$\pm$1.30} & 96.96$\pm$1.43 & \underline{96.99$\pm$1.51} & 96.93$\pm$1.34 & 96.70$\pm$1.16 & \textbf{97.13$\pm$1.33} &  & \textbf{97.25$\pm$1.30} & \textbf{97.25$\pm$1.52} &  & \textbf{97.37$\pm$1.49 }& \textbf{97.37$\pm$0.94}
\\
Wine & 95.28$\pm$2.36 &  & 97.31$\pm$1.94 & 96.67$\pm$1.76 & 96.85$\pm$2.26 & 97.41$\pm$1.84 & \textbf{97.69$\pm$1.79} & \textbf{97.69$\pm$1.89} &  & \textbf{97.69$\pm$1.79} & 97.22$\pm$2.04 &  & 97.04$\pm$2.71 & \textbf{97.22$\pm$1.95}
\\
Yeast* & 70.33$\pm$10.50 &  & 69.84$\pm$10.26 & 70.26$\pm$10.51 & 70.29$\pm$10.52 & 69.86$\pm$10.29 & \textbf{70.32$\pm$10.51} & \textbf{70.32$\pm$10.51} &  & 68.81$\pm$11.35 & \textbf{69.90$\pm$10.35} &  & 66.01$\pm$13.21 & \textbf{69.41$\pm$10.36}
\\
Credit & 76.40 &  & 76.41 & 76.68 & \textbf{76.96} & 76.50 & 76.87 & \underline{76.89}&  & \textbf{76.45} & 76.29 &  & \textbf{81.15} & 81.07\\
\midrule
\# outperform & - &  & 12 & 11 & 12 & 12 & 12 & - &  & 9 & - &  & 7 & -\\
\bottomrule
\end{tabular}
\textit{For methods with LMNN as the backbone, the best ones are shown in bold and the second best ones are underlined; for methods with SCML or PNCA as the backbone, the best ones are shown in bold. `\# outperform' counts the number of datasets where LMNN-PL (SCML-PL, PNCA-PL resp.) outperforms or performs equally well with LMNN-based (SCML, PNCA resp.) methods.}
\end{threeparttable}

\vspace{0.25in}
\caption{Classification accuracy (or G-means indicated by an asterisk) of 3NN noise-contaminated datasets. Gaussian noise with an SNR of 5~dB is added to test data.}
\label{tab:noise_spherical}
\centering
\small
\setlength\tabcolsep{2pt}
\begin{tabular}{lccc cccc cccc ccc}
\toprule
&&& \multicolumn{6}{c}{LMNN-based}&& \multicolumn{2}{c}{SCML-based}&& \multicolumn{2}{c}{PNCA-based}\\
\cmidrule{4-9} \cmidrule{11-12} \cmidrule{14-15}
Dataset & Euclidean &  & LMNN & LDD & CAP & SN & DRIFT & LMNN-PL &  & SCML & SCML-PL &  & PNCA & PNCA-PL\\
\midrule
Australian & 82.28$\pm$1.67 &  & 82.46$\pm$1.58 & \underline{83.02$\pm$1.58} & 82.36$\pm$1.56 & 82.56$\pm$1.54 & 82.58$\pm$1.45 & \textbf{83.50$\pm$1.56} &  & 82.93$\pm$1.65 & \textbf{83.42$\pm$1.68} &  & 82.79$\pm$2.37 & \textbf{83.66$\pm$2.28}
\\
Breast cancer & 96.79$\pm$1.05 &  & 96.25$\pm$1.09 & \underline{96.69$\pm$1.09} & 96.35$\pm$1.02 & 96.29$\pm$1.11 & 96.66$\pm$1.00 & \textbf{96.71$\pm$1.08} &  & 96.40$\pm$1.05 & \textbf{96.65$\pm$1.06} &  & 96.20$\pm$1.37 & \textbf{96.77$\pm$1.14}
\\
Ecoli* & 79.95$\pm$7.67 &  & 74.96$\pm$7.16 & \underline{79.13$\pm$7.67} & 74.46$\pm$9.36 & 75.19$\pm$7.23 & 77.86$\pm$7.76 & \textbf{80.04$\pm$7.51} &  & 76.49$\pm$6.78 & \textbf{78.38$\pm$7.92} &  & 75.52$\pm$6.08 & \textbf{77.04$\pm$6.20}
\\
Fourclass & 69.11$\pm$1.12 &  & 67.62$\pm$1.23 & 68.77$\pm$1.14 & 67.63$\pm$1.12 & 68.55$\pm$1.30 & \textbf{69.03$\pm$1.13} & \underline{69.01$\pm$1.17} &  & 68.07$\pm$1.16 & \textbf{68.86$\pm$1.06} &  & \textbf{70.42$\pm$2.17} & 69.39$\pm$4.60
\\
Haberman & 69.93$\pm$1.88 &  & 69.84$\pm$1.79 & \textbf{69.92$\pm$1.87} & 69.23$\pm$2.00 & \underline{69.90$\pm$1.88} & 69.09$\pm$2.49 & 69.89$\pm$1.90 &  & 69.65$\pm$1.63 & \textbf{69.88$\pm$1.83} &  & \textbf{74.32$\pm$3.22} & \textbf{74.32$\pm$3.10}
\\
Iris & 79.75$\pm$3.26 &  & 78.61$\pm$2.97 & \underline{78.87$\pm$3.16} & 77.79$\pm$3.27 & 78.70$\pm$3.08 & 78.43$\pm$3.09 & \textbf{79.04$\pm$3.09} &  & 78.16$\pm$3.58 & \textbf{79.01$\pm$3.12 }&  & 77.95$\pm$4.53 & \textbf{78.20$\pm$3.90}
\\
Segment & 88.18$\pm$0.64 &  & 81.02$\pm$3.55 & \underline{86.15$\pm$1.26} & 85.34$\pm$2.47 & 82.10$\pm$3.41 & \textbf{86.63$\pm$1.09} & 84.72$\pm$2.62 &  & 60.18$\pm$9.73 & \textbf{61.33$\pm$9.05} &  & 78.27$\pm$2.83 & \textbf{80.28$\pm$3.35}
\\
Sonar & 83.47$\pm$3.21 &  & 83.56$\pm$4.27 & \textbf{86.18$\pm$2.95} & \underline{85.41$\pm$2.82} & 83.52$\pm$4.28 & 84.65$\pm$3.30 & 85.00$\pm$3.15 &  & 77.01$\pm$4.23 & \textbf{79.49$\pm$3.80} &  & 80.74$\pm$4.36 & \textbf{81.74$\pm$3.43}
\\
Voting & 93.19$\pm$1.15 &  & 94.00$\pm$1.00 & 94.25$\pm$1.14 & \underline{94.37$\pm$1.17} & 94.06$\pm$1.00 & 93.95$\pm$1.12 & \textbf{94.64$\pm$1.21} &  & 93.99$\pm$1.15 & \textbf{94.64$\pm$1.09} &  & 92.61$\pm$1.64 & \textbf{93.46$\pm$1.77}
\\
WDBC & 95.92$\pm$1.30 &  & 91.71$\pm$1.90 & \textbf{96.30$\pm$0.94} & \underline{96.16$\pm$1.08} & 92.46$\pm$1.80 & 96.04$\pm$0.86 & 96.11$\pm$0.88 &  & 95.74$\pm$1.30 & \textbf{96.21$\pm$1.16} &  & 96.03$\pm$1.54 & \textbf{96.22$\pm$1.15}
\\
Wine & 94.20$\pm$1.46 &  & 93.33$\pm$1.63 & 94.03$\pm$1.39 & 93.97$\pm$1.47 & 93.45$\pm$1.70 & \textbf{94.66$\pm$1.15} & \underline{94.51$\pm$1.20} &  & 94.01$\pm$1.56 & \textbf{94.61$\pm$1.32} &  & \textbf{94.19$\pm$1.94} & 93.48$\pm$1.48
\\
Yeast* & 69.36$\pm$10.47 &  & 54.13$\pm$8.24 & 68.62$\pm$10.43 & 66.48$\pm$10.18 & 55.49$\pm$9.45 & \underline{69.64$\pm$10.49} & \textbf{69.82$\pm$10.44} &  & 55.96$\pm$7.64 & \textbf{60.47$\pm$10.39} &  & 61.41$\pm$17.97 & \textbf{63.59$\pm$18.33}
\\
Credit & 76.28 &  & 76.16 & 76.22 & 76.05 & \underline{76.30} & \textbf{76.37} & 76.15 &  & \textbf{75.93} & 75.55 &  & 78.24 & \textbf{79.13}\\
\midrule
\# outperform & - &  & 12 & 9 & 10 & 11 & 9 & - &  & 12 & - &  & 11 & -\\
\bottomrule
\end{tabular}
\end{sidewaystable}

\subsubsection{Investigation into robustness}
\label{subsec:exp_robustness}

To test robustness, we add zero-mean Gaussian noise with a diagonal covariance matrix and equal variances to test data; the noise intensity is controlled via the signal-to-noise ratio (SNR) and chosen as 5~dB. In addition, considering the small sample size of UCI datasets, we augment test data by adding multiple rounds of random noise until its size reaches 10,000. As shown in Table~\ref{tab:noise_spherical}, the proposed methods achieve higher classification accuracy or G-means than the corresponding baselines on almost all datasets. Moreover, LMNN-PL is superior to existing regularization techniques or robust metric learning methods on at least 9 datasets. These results clearly demonstrate the efficacy of adding perturbation loss for improving robustness against instance perturbation. Additional experiments with other noise types and intensities are reported in Section~\ref{sec:additional_exp} in the Supplementary Material, where we observe similar advantages of the proposed loss. 

\clearpage
\subsection{Experiments on high-dimensional data}
\label{subsec:exp_HD}

As mentioned in Remark 3, we extend LMNN-PL for high-dimensional data with PCA being used as a pre-processing step. To verify its effectiveness, we test it on the following three datasets: 
\begin{enumerate}
    \item Isolet~\cite{Dua:2019}: The dataset is a spoken letter database and is available from UCI. It includes 7,797 instances, grouped into four training sets and one test set. We apply PCA to reduce the feature dimension from 617 to 170, accounting for 95\% of total variance. All methods are trained four times, one time on each training set, and evaluated on the pre-given test set.
    \item MNIST-2k~\cite{cai2010graph}: The dataset includes the first 2,000 training images and first 2,000 test images of the MNIST database. PCA is applied to reduce the dimension from 784 to 141, retaining 95\% of total variance. All methods are trained and tested once on the pre-given training/test partition. 
    \item APS Failure~\cite{Dua:2019}: This is a multivariate dataset with a highly imbalanced class distribution. The training set includes 60,000 instances, among which 1,000 belong to the positive class. The test set includes 16,000 instances with 375 positive ones. The training set is further split into 40,000 instances for training and 20,000 instances for selecting hyperparameters. All methods are tested once on the test set. Applying PCA reduces the feature dimension from 161 to 79. Due to the large sample size, we only evaluate LMNN and LMNN-PL on this dataset. 
\end{enumerate}

In addition to aforementioned methods, we introduce CAP-PL, which comprises the triplet loss of LMNN, the regularizer of CAP, and the proposed perturbation loss. CAP enforces $\bm M$ to be low-rank, which is a suitable constraint for high-dimensional data. With the inclusion of perturbation loss, we expect the learned compact metric to be more robust to perturbation. For a fair comparison, in CAP-PL, we use the same rank and regularization weight as CAP, and tune $\tau,\lambda$ from 10 randomly sampled sets of values. 

\begin{table}[t]
\caption{Generalization and robustness of selected metric learning methods on high-dimensional datasets.}
\label{tab:HD}
\small
\centering
\setlength\tabcolsep{1.5pt}
\begin{threeparttable}
\begin{tabular}{lcccccc}
\toprule
\multicolumn{7}{c}{Isolet }\\
\midrule
Method & Clean & IG,SNR=20 & IG,SNR=5 & AG,SNR=20 & AG,SNR=5 & Adv. \\
 &  & (0.0809) & (0.4233) & (0.0588) & (0.3181) & margin \\
\midrule
LMNN & 90.14$\pm$4.45 & 90.09$\pm$4.15 & 86.02$\pm$3.48 & 90.17$\pm$4.03 & 87.81$\pm$3.87 & 0.1095\\
LMNN-PL & 91.08$\pm$3.71 & 91.02$\pm$3.77 & 87.91$\pm$3.30 & 91.05$\pm$3.73 & 89.40$\pm$3.76 & 0.1249\\
\midrule
SCML & 90.73$\pm$4.10 & 90.33$\pm$4.21 & 86.50$\pm$4.18 & 90.51$\pm$4.14 & 88.50$\pm$3.71 & 0.0683\\
SCML-PL & 90.83$\pm$4.16 & 90.67$\pm$4.12 & 86.55$\pm$3.75 & 90.83$\pm$4.16 & 88.41$\pm$4.07 & 0.0822\\
\midrule
CAP & 91.05$\pm$3.66 & 91.13$\pm$3.85 & 88.97$\pm$4.00 & 91.10$\pm$3.73 & 89.90$\pm$3.87 & 0.1514\\
CAP-PL & 91.58$\pm$3.96 & 91.52$\pm$3.86 & 89.91$\pm$3.74 & 91.47$\pm$3.91 & 90.65$\pm$3.73 & 0.1559\\
\bottomrule
\toprule
\multicolumn{7}{c}{MNIST }\\
\midrule
Method & Clean & IG,SNR=20 & IG,SNR=5 & AG,SNR=20 & AG,SNR=5 & Adv.\\
&  & (0.0540) & (0.2939) & (0.0649) & (0.3482) & margin\\
\midrule
LMNN & 90.55 & 90.00 & 88.40 & 90.10 & 88.40 & 0.1528\\
LMNN-PL & 91.15 & 91.35 & 90.80 & 91.45 & 90.35 & 0.2235\\
\midrule
SCML & 88.95 & 88.75 & 87.35 & 88.85 & 86.45 & 0.1217\\
SCML-PL & 89.15 & 89.20 & 88.50 & 89.35 & 88.05 & 0.1432\\
\midrule
CAP & 91.65 & 91.80 & 91.40 & 91.80 & 90.70 & 0.2219\\
CAP-PL & 92.00 & 91.90 & 90.85 & 91.95 & 90.65 & 0.2264\\
\bottomrule
\toprule
\multicolumn{7}{c}{APS Failure}\\
\midrule
Method & Clean & IG,SNR=20 & IG,SNR=5 & AG,SNR=20 & AG,SNR=5 & Adv.\\
 &  & (0.0906) & (0.5097) & (0.0996) & (0.5604) & margin\\
\midrule
LMNN & 80.69 & 80.20 & 75.66 & 81.18 & 75.13 & 0.1773\\
LMNN-PL & 80.89 & 82.15 & 74.13 & 82.33 & 77.92 & 0.2583\\
\bottomrule
\end{tabular}
\smallskip
\textit{Columns 3-6 report methods' robustness against isotropic Gaussian noise (IG) and anisotropic Gaussian noise (AG). Values in brackets give the average perturbation size, calculated as the mean value of the $L_2$-norm of noises ($\|\Delta \bm x_i\|_2$).}
\end{threeparttable}
\end{table}

Table~\ref{tab:HD} compares the generalization and robustness performance of LMNN, CAP, SCML, and our method; the generalization performance of other methods are inferior to LMNN-PL and reported in Table~\ref{tab:HD_additional} in the Supplementary Material. First, on all three original datasets, our method achieves better performance than the baseline methods, validating its efficacy in improving the generalization ability of the learned metric. Secondly, when the SNR is 20~dB, the average perturbation size is smaller than the average adversarial margin. In this case, our method maintains its superiority. When the SNR is 5~dB, the average perturbation size is larger than the average adversarial margin. Nonetheless, our method produces even larger gain in classification performance for LMNN on all datasets except APS Failure with the Gaussian noise, for SCML on MNIST and on Isolet with the Gaussian noise, for CAP on Isolet. These results suggest that adversarial margin is indeed a contributing factor in enhancing robustness. Thirdly, CAP-PL obtains higher accuracy on both clean and noise-contaminated data than LMNN-PL. This supports our discussion in  Section~\ref{sec:related_work} that regularization and perturbation loss impose different requirements on $\bm M$ and combining them has the potential for learning a more effective distance metric.

\subsection{Computational cost}
We now analyze the computational complexity of LMNN-PL. According to Eq.~\ref{eq:grad}, our method requires additional calculations on $d^2_{\bm M^2}(\bm x_j, \bm x_l)$ and $\bm M \bm X_{jl}$. Given $n$ training instances, $k$ target neighbors and $p$ features, the computational complexities of $d^2_{\bm M^2}(\bm x_j, \bm x_l)$ and $\bm M \bm X_{jl}$ are $O(np^2+n^2p)$ and $O(n^2p^2)$, respectively. The total complexity is $O(p^3+n^2p^2+kn^2p)$, same as that of LMNN.

Table~\ref{tab:training_time} compares the running time of LMNN-based methods on five UCI datasets that are large in sample size or in dimensionality and two high-dimensional datasets. The computational cost of our method is comparable to LMNN. 

\begin{table}[t]
\caption{Average training time (in seconds) of LMNN-based methods.}
\label{tab:training_time}
\centering
\small
\begin{tabular}{lrrrrrr}
\toprule
 & LMNN & LDD & CAP & SN & DRIFT & LMNN-PL\\
\midrule
Australian & 13.44 & 0.83 & 3.07 & 7.60 & 1.00 & 2.15\\
Segment & 27.48 & 10.45 & 11.47 & 24.66 & 5.12 & 19.54\\
Sonar & 4.93 & 4.08 & 4.65 & 30.39 & 0.92 & 6.75\\
WDBC & 9.38 & 2.94 & 5.22 & 16.54 & 5.12 & 8.17\\
Credit & 724.42 & 34.65 & 115.22 & 966.63 & 130.36 & 138.63\\
Isolet & 339.57 & 207.69 & 176.50 & 540.26 & NA & 190.55\\
MNIST & 369.55 & 68.98 & 289.18
 & 197.50 & 37.51 & 391.04\\
\bottomrule
\end{tabular}
\vspace{1em}
\end{table}

\begin{figure*}[t]
    \centering
    {\includegraphics[width=.325\linewidth]{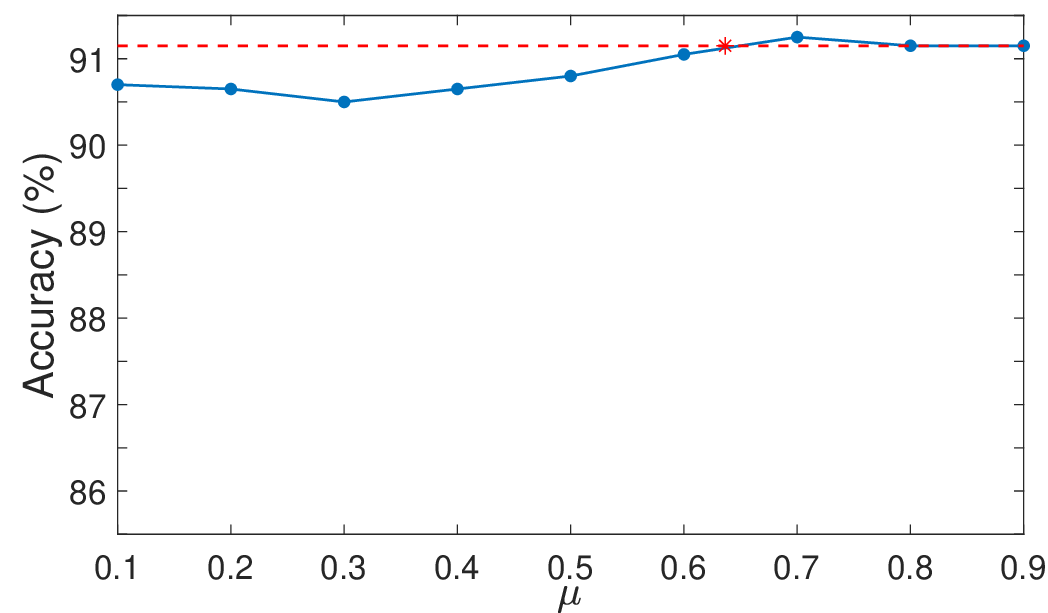}}\hfill%
    {\includegraphics[width=.325\linewidth]{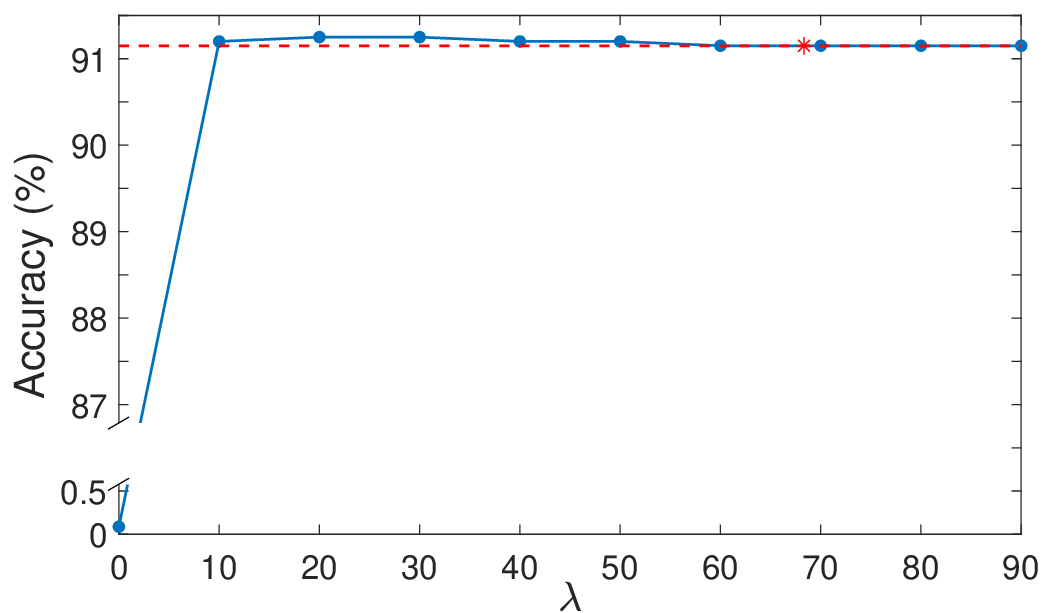}}\hfill%
    {\includegraphics[width=.325\linewidth]{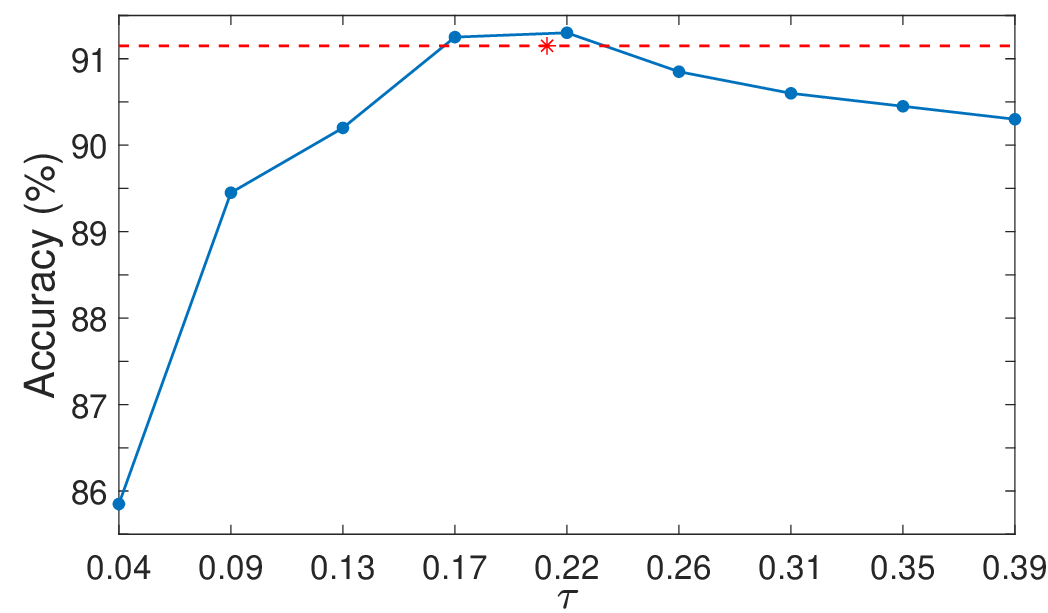}}\hfill
    \caption{Sensitivity of LMNN-PL to hyperparameters (indicated by the straight line). The optimal accuracy and parameter value found via CV are indicated by the dashed line and asterisk, respectively. }
    \label{fig:parameter_sensitivity}
\end{figure*}

\subsection{Parameter sensitivity}
\label{subsec:hyper_sensitivity}

The proposed LMNN-PL includes three hyperparameters -- $\mu$ for the weight of similarity constraints, $\lambda$ for the weight of the perturbation loss, and $\tau$ for the desired adversarial margin. We investigate their influences on the classification
performance by varying one hyperparameter and fixing the
other two at their optimal values. Fig.~\ref{fig:parameter_sensitivity} shows the accuracy on MNIST evaluated over the range of the hyperparameter. The performance changes smoothly with respect to $\mu$. It is stable over a wide range of $\lambda$. When $\lambda$ equals 0, LMNN-PL fails to learn a metric and returns a zero matrix. The performance is most affected by $\tau$. Indeed, $\tau$ plays the central role in LMNN-PL as it determines the distribution of adversarial margins. A small value of $\tau$ has little influence on the objective function as the adversarial margin of most instances may already exceed it before optimization, and a large value may greatly reduce the number of triplets that satisfy the loss condition in the definition of pseudo-robustness (i.e., $\hat{n}(t_{\bm s})$ in Theorem~\ref{thm:bound}). Therefore, we shall strive to search for its optimal value.

\section{Conclusion}
In this paper, we propose to enhance the robustness and generalization of distance metrics. This is easily achievable by taking advantage of the linear transformation induced by the Mahalanobis distance. Specifically, we find an explicit formula for the adversarial margin, which is defined as the Euclidean distance between benign instances and their closest adversarial examples, and advocate to enlarge it through penalizing the perturbation loss designed on the basis of the derivation. Experiments verify that our method effectively enlarges the adversarial margin, sustains classification excellence, and enhances robustness to instance perturbation. The proposed loss term is generic in nature and could be readily embedded in other Mahalanobis-based metric learning methods. In the future, we will consider extending the idea to metric learning methods with nonlinear feature extraction and/or nonlinear metric learning methods.

\bibliographystyle{abbrv}
\bibliography{reference}

\newpage
\appendix

\section{Derivation of the closest adversarial example, adversarial margin and gradient of perturbation loss}
\label{sec:derivation}

In the following derivations, we consider heterogeneous and correlated perturbations; the homogeneous and uncorrelated perturbations, discussed in the main text, can be seen as a special case. 

First, we characterize heterogeneous and correlated perturbations through hyperellipsoids. In terms of the quadratic form, an arbitrarily oriented hyperellipsoid, centered at $\bm \mu \in \mathbb{R}^p$, is defined by the solutions to the equation 
\begin{equation*}
\footnotesize
\begin{split}
    \{\bm x \in \mathbb{R}^p: (\bm x-\bm \mu)^T \bm A_0 (\bm x-\bm \mu) = r^2\},
\end{split}
\end{equation*}
where $\bm A_0$ is a positive definite matrix. By the Cholesky decomposition, $\bm A_0 = \bm A \bm A^T$. Therefore, finding the closest adversarial example of $\bm x_i$ on the hyperellipsoid is equivalent to finding the point $\bm x'_i$ that defines the smallest hypersphere given by $(\bm A^T (\bm x'_i-\bm x_i))^T (\bm A^T (\bm x'_i-\bm x_i)) = r^2$.

The closest adversarial example is identified by optimizing Eq.~\ref{eq:x_min_opt(app)}; its equivalent form is given in Eq.~\ref{eq:x_min_opt_equivalent}:
\begin{equation}
\footnotesize
\begin{split}
    &\bm x_{i,{\rm min}}=\arg\min_{\bm x'_i\in \mathbb{R}^p} (\bm x'_i - \bm x_i)^T \bm A_0 (\bm x'_i - \bm x_i)\\ 
    &\text{s.t. } (\bm L \bm x'_i - \frac{\bm L \bm x_j + \bm L \bm x_l}{2})^T (\bm L \bm x_l - \bm L \bm x_j) = 0.
\end{split}
\label{eq:x_min_opt(app)}
\end{equation}
\vspace{-1em}
\begin{equation}
\footnotesize
\begin{split}
    &\bm x_{i,{\rm min}}=\arg\min_{\bm x'_i\in \mathbb{R}^p} \big(\bm A^T (\bm x'_i-\bm x_i)\big)^T \big(\bm A^T (\bm x'_i-\bm x_i)\big) \\ 
    &\text{s.t. } (\bm L \bm x'_i - \frac{\bm L \bm x_j + \bm L \bm x_l}{2})^T (\bm L \bm x_l - \bm L \bm x_j) = 0.
\label{eq:x_min_opt_equivalent}
\end{split}
\end{equation}
Applying the method of Lagrangian multiplier, we transform the above problem to the following Lagrangian function by introducing the Lagrangian multiplier $\lambda$ and then solve it by setting the first partial derivatives to zero: 
\begin{equation*}
\scriptsize
\begin{split}
    &\min_{\bm x'_i} \big(\bm A^T (\bm x'_i-\bm x_i)\big)^T \big(\bm A^T (\bm x'_i-\bm x_i)\big) - \lambda (\bm L \bm x'_i - \frac{\bm L \bm x_j + \bm L \bm x_l}{2})^T (\bm L \bm x_l - \bm L \bm x_j)
\end{split}
\end{equation*}
\begin{equation*}
\scriptsize
\begin{split}
    &\frac{\delta}{\delta \bm x'_i}: 2 \bm A \bm A^T (\bm x'_i - \bm x_i) - \lambda \bm L^T \bm L(\bm x_l - \bm x_j) = 0 \\
    &\hspace{2.2em} \bm x'_i= \bm x_i + \frac{\lambda}{2} \bm A_0^{-1} \bm L^T \bm L (\bm x_l - \bm x_j) \\
    & \Big(\bm L \bm x_i + \frac{\lambda}{2} \bm L \bm A_0^{-1} \bm L^T \bm L (\bm x_l - \bm x_j) - \frac{\bm L \bm x_j + \bm L \bm x_l}{2} \Big)^T (\bm L \bm x_l - \bm L \bm x_j) = 0 \\
    &\hspace{2.2em} \frac{\lambda}{2} = \frac{(\frac{\bm x_j + \bm x_l}{2}-\bm x_i)^T \bm L^T \bm L (\bm x_l - \bm x_j)}{(\bm x_l - \bm x_j)^T \bm L^T \bm L \bm A_0^{-1} \bm L^T \bm L (\bm x_l - \bm x_j)}\\
    &\hspace{0.5em} \bm x_{i,{\rm min}} = \bm x_i + \frac{(\frac{\bm x_j + \bm x_l}{2} - \bm x_i)^T \bm M (\bm x_l - \bm x_j)}{(\bm x_l - \bm x_j)^T \bm M \bm A_0^{-1} \bm M (\bm x_l - \bm x_j)} \bm A_0^{-1} \bm M (\bm x_l - \bm x_j)\\
\end{split}
\end{equation*}
The Hessian matrix equals $2\bm A_0$, which is positive definite, and hence $\bm x_{i,{\rm min}}$ is the minimum point. Replacing $\bm A_0 = \bm I$ (identity matrix) gives Eq.~\ref{eq:x_min_def}.

Next, we calculate the squared adversarial margin by first simplifying $\bm x_{i,{\rm min}}$ and then computing $r^2$ as follows: 
\begin{equation*}
\footnotesize
\begin{split}
    &(\frac{\bm x_j + \bm x_l}{2} - \bm x_i)^T \bm M (\bm x_l - \bm x_j) \\
    =& \frac{1}{2} [(\bm x_j - \bm x_i) + (\bm x_l - \bm x_i)]^T \bm M [(\bm x_l - \bm x_i) - (\bm x_j - \bm x_i) ]\\
    =& \frac{1}{2}[d^2_{\bm M} (\bm x_i, \bm x_l) - d^2_{\bm M} (\bm x_i, \bm x_j)]\\
    r^2 =& (\bm x_i - \bm x_{i,{\rm min}})^T \bm A_0 (\bm x_i - \bm x_{i,{\rm min}}) \\
    =&\left( \frac{d^2_{\bm M} (\bm x_i, \bm x_l) - d^2_{\bm M} (\bm x_i, \bm x_j)}{2 (\bm x_l - \bm x_j)^T \bm M \bm A_0^{-1} \bm M (\bm x_l - \bm x_j)}\bm A_0^{-1} \bm L^T \bm L (\bm x_l - \bm x_j) \right)^T \\
    & \bm A_0 \left( \frac{d^2_{\bm M} (\bm x_i, \bm x_l) - d^2_{\bm M} (\bm x_i, \bm x_j)}{2 (\bm x_l - \bm x_j)^T \bm M \bm A_0^{-1} \bm M (\bm x_l - \bm x_j)}\bm A_0^{-1} \bm L^T \bm L (\bm x_l - \bm x_j) \right) \\
    =&\frac{\left(d^2_{\bm M} (\bm x_i, \bm x_l) - d^2_{\bm M} (\bm x_i, \bm x_j)\right)^2}{4 \left((\bm x_l - \bm x_j)^T \bm M \bm A_0^{-1} \bm M (\bm x_l - \bm x_j)\right)^2} \cdot \\
    &(\bm x_l - \bm x_j)^T \bm L^T \bm L \bm A_0^{-1} \bm A_0 \bm A_0^{-1} \bm L^T \bm L (\bm x_l - \bm x_j) \\
    =&\frac{\left(d^2_{\bm M} (\bm x_i, \bm x_l) - d^2_{\bm M} (\bm x_i, \bm x_j)\right)^2}{4 \left((\bm x_l - \bm x_j)^T \bm M \bm A_0^{-1} \bm M (\bm x_l - \bm x_j)\right)}
\end{split}
\end{equation*}
Substituting $\bm A_0=\bm I$ gives Eq.~\ref{eq:adv_margin}.

Finally, we derive the gradient of $J_\text{P}$ with respect to $\bm M$. When $d^2_{\bm M}(\bm x_i, \bm x_l) > d^2_{\bm M}(\bm x_i, \bm x_j)$ and $r \geq \tau$ (i.e. $\tilde{d}_{\bm E}(\bm x_i, \bm x_{i,{\rm min}}) \geq \tau$ in the hypersphere case), or $d^2_{\bm M}(\bm x_i, \bm x_l) \leq d^2_{\bm M}(\bm x_i, \bm x_j)$, the gradient equals zero. When $d^2_{\bm M}(\bm x_i, \bm x_l) > d^2_{\bm M}(\bm x_i, \bm x_j)$ and $r < \tau$, the gradient of $J_\text{P}$ equals the gradient of $-r^2$ (i.e. $-\tilde{d}_{\bm E}^2(\bm x_i, \bm x_{i,{\rm min}})$ in the hypersphere case), which can be calculated by using the quotient rule and the derivative of trace~\cite{petersen2012matrix}: 
\begin{equation*}
\footnotesize
\begin{split}
    \frac{\partial}{\partial \bm M}& \left(d^2_{\bm M}(\bm x_i, \bm x_l) - d^2_{\bm M}(\bm x_i, \bm x_j) \right)^2 \\
    =& 2\left(d^2_{\bm M}(\bm x_i, \bm x_l) - d^2_{\bm M}(\bm x_i, \bm x_j)\right) (\bm X_{il}-\bm X_{ij})\\
    \frac{\partial}{\partial \bm M} &(\bm x_l - \bm x_j)^T \bm M \bm A_0^{-1} \bm M (\bm x_l - \bm x_j) 
    = \frac{\partial}{\partial \bm M} \tr (\bm X_{jl} \bm M \bm A_0^{-1} \bm M) \\
    =&\bm X_{jl} \bm M \bm A_0^{-1} +\bm A_0^{-1} \bm M \bm X_{jl}\\
    \frac{\partial}{\partial \bm M} &\frac{\left(d^2_{\bm M} (\bm x_i, \bm x_l) - d^2_{\bm M} (\bm x_i, \bm x_j)\right)^2}{4 \left((\bm x_l - \bm x_j)^T \bm M \bm A_0^{-1} \bm M (\bm x_l - \bm x_j)+\epsilon\right)} \\
    =& \frac{2\left(d^2_{\bm M}(\bm x_i, \bm x_l) - d^2_{\bm M}(\bm x_i, \bm x_j)\right) (\bm X_{il}-\bm X_{ij})}{4\left((\bm x_l - \bm x_j)^T \bm M \bm A_0^{-1} \bm M (\bm x_l - \bm x_j)+\epsilon\right)} \\
    -&\frac{\left(d^2_{\bm M}(\bm x_i, \bm x_l) - d^2_{\bm M}(\bm x_i, \bm x_j)\right)^2 (\bm X_{jl} \bm M \bm A_0^{-1} +\bm A_0^{-1} \bm M \bm X_{jl})}{4\left((\bm x_l - \bm x_j)^T \bm M \bm A_0^{-1} \bm M (\bm x_l - \bm x_j)+\epsilon\right)^2},
\end{split}
\end{equation*}
where $\tr(\cdot)$ denotes the trace operator. $\bm X_{ij} = (\bm x_i-\bm x_j) (\bm x_i-\bm x_j)^T$ and $\bm X_{il},\bm X_{jl}$ are defined similarly. Substituting $\bm A_0=\bm I$ gives Eq.~\ref{eq:grad}.

\subsection{Derivation of closest adversarial example, adversarial margin, and gradient of perturbation loss in the high-dimensional case}
\label{sec:derivation_HD}

The closet adversarial example, adversarial margin and gradient of the perturbation loss with dimensionality reduction are derived by following the same principle as in Appendix~\ref{sec:derivation}. 

The method of Lagrangian multiplier is applied to derive a closed-form solution to the closest adversarial example:
\begin{equation*}
\footnotesize
\begin{split}
    \min_{\bm x_i'} \big(\bm A^T (&\bm x'_i-\bm x_i)\big)^T \big(\bm A^T (\bm x'_i-\bm x_i)\big) \\
    - \lambda (&\bm x_i' - \frac{\bm x_j + \bm x_l}{2})^T \bm D^T \bm L^T \bm L \bm D (\bm x_l - \bm x_j) \\
    \frac{\delta}{\delta \bm x'_i}: \bm x'_i =& \bm x_i + \frac{\lambda}{2} \bm A_0^{-1} \bm D^T \bm L^T \bm L \bm D(\bm x_l - \bm x_j) \\
    \frac{\lambda}{2} =& \frac{(\frac{\bm x_j + \bm x_l}{2}-\bm x_i)^T \bm D^T \bm L^T \bm L \bm D (\bm x_l - \bm x_j)}{(\bm x_l - \bm x_j)^T \bm D^T \bm L^T \bm L \bm D \bm A_0^{-1} \bm D^T \bm L^T \bm L \bm D(\bm x_l - \bm x_j)}\\
    \bm x_{i,{\rm min}}^\text{PCA} 
    =& \bm x_i + \frac{(\frac{\tilde{\bm x}_j +\tilde{\bm x}_l}{2} - \tilde{\bm x}_i)^T \bm M (\tilde{\bm x}_l - \tilde{\bm x}_j)}{(\tilde{\bm x}_l - \tilde{\bm x}_j)^T \bm M  \bm D \bm A_0^{-1} \bm D^T \bm M \bm (\tilde{\bm x}_l - \tilde{\bm x}_j)} \cdot \\
    &\bm A_0^{-1} \bm D^T \bm M (\tilde{\bm x}_l - \tilde{\bm x}_j),
\end{split}
\end{equation*}
where $\tilde{\bm x}$ denotes $\bm D \bm x$.

The squared adversarial margin is calculated from the definition of the hyperellipsoid: 
\begingroup
\begin{equation*}
\footnotesize
\begin{split}
    r^2 =&(\bm x_i-\bm x_{\rm min}^\text{PCA})^T \bm A_0 (\bm x_i-\bm x_{\rm min}^\text{PCA}) \\
    =&\frac{\big(d^2_{\bm M} (\tilde{\bm x}_i, \tilde{\bm x}_l) - d^2_{\bm M} (\tilde{\bm x}_i, \tilde{\bm x}_j)\big)^2}{4 \big( (\tilde{\bm x}_l - \tilde{\bm x}_j)^T \bm L^T \bm L \bm D \bm A_0^{-1} \bm D^T \bm L^T \bm L(\tilde{\bm x}_l - \tilde{\bm x}_j) \big)^2}\\
    &\cdot(\tilde{\bm x}_l - \tilde{\bm x}_j)^T \bm L^T \bm L \bm D \bm A_0^{-1} \bm A_0 \bm A_0^{-1} \bm D^T \bm L^T \bm L (\tilde{\bm x}_l - \tilde{\bm x}_j)  \\
    =& \frac{\big(d^2_{\bm M}(\tilde{\bm x}_i,\tilde{\bm x}_l)-d^2_{\bm M}(\tilde{\bm x}_i,\tilde{\bm x}_j)\big)^2}{4(\tilde{\bm x}_l - \tilde{\bm x}_j)^T \bm M \bm D \bm A_0^{-1} \bm D^T \bm M (\tilde{\bm x}_l-\tilde{\bm x}_j)}
\end{split}
\end{equation*}

The perturbation loss is defined similarly to Eq.~4 as follows:
\begin{equation*}
\footnotesize
\begin{split}
    J_\text{P}^\text{PCA} =& \frac{1}{|\mathcal{R}|}\sum_\mathcal{R}
    \big \{ [\tau^2 - \tilde{d}_{\bm E}^2(\bm x_i, \bm x^\text{PCA}_{i,{\rm min}})]_+ \mathds{1}_{\{d^2_{\bm M}(\tilde{\bm x}_i, \tilde{\bm x}_l) > d^2_{\bm M}(\tilde{\bm x}_i, \tilde{\bm x}_j)\}}\\
    +&
    \tau^2 \mathds{1}[d^2_{\bm M}(\tilde{\bm x}_i, \tilde{\bm x}_l) \leq d^2_{\bm M}(\tilde{\bm x}_i, \tilde{\bm x}_j)]\big\} ,
\end{split}
\end{equation*}
where $\tilde{d}_{\bm E}^2(\bm x_i, \bm x^\text{PCA}_{i,{\rm min}})=\frac{\big(d^2_{\bm M}(\tilde{\bm x}_i,\tilde{\bm x}_l)-d^2_{\bm M}(\tilde{\bm x}_i,\tilde{\bm x}_j)\big)^2}{4\left((\tilde{\bm x}_l - \tilde{\bm x}_j)^T \bm M \bm D \bm D^T \bm M (\tilde{\bm x}_l-\tilde{\bm x}_j)+\epsilon\right)}$.

The gradient of $J_\text{P}^\text{PCA}$ is given as: 
\begin{equation*}
\scriptsize
\begin{split}
    \frac{\partial J_\text{P}^\text{PCA}}{\partial \bm M}
    =& \frac{1}{|\mathcal{R}|} \sum_\mathcal{R} \alpha_{ijl} \bigg\{ \frac{\big(d^2_{\bm M}(\tilde{\bm x}_i,\tilde{\bm x}_l)-d^2_{\bm M}(\tilde{\bm x}_i,\tilde{\bm x}_j)\big) (\tilde{\bm X}_{ij}-\tilde{\bm X}_{il})}{2\big((\tilde{\bm x}_l - \tilde{\bm x}_j)^T \bm M \bm D \bm A_0^{-1} \bm D^T \bm M (\tilde{\bm x}_l-\tilde{\bm x}_j)+\epsilon\big)} \\
    +&\frac{\big(d^2_{\bm M}(\tilde{\bm x}_i,\tilde{\bm x}_l)-d^2_{\bm M}(\tilde{\bm x}_i,\tilde{\bm x}_j)\big)^2}{4\big((\tilde{\bm x}_l - \tilde{\bm x}_j)^T \bm M \bm D \bm A_0^{-1} \bm D^T \bm M (\tilde{\bm x}_l-\tilde{\bm x}_j)+\epsilon\big)^2} \\
    &\cdot(\tilde{\bm X}_{jl} \bm M \bm D \bm A_0^{-1} \bm D^T + \bm D \bm A_0^{-1} \bm D^T \bm M \tilde{\bm X}_{jl})\bigg\}.
\end{split}
\end{equation*}

\section{Additional illustrations of robustifying metric learning methods}
\label{sec:CR_extension}

In this section, we incorporate the proposed perturbation loss into two triplet-based metric learning methods, sparse compositional metric learning (SCML) and proxy neighborhood component analysis (PNCA). 

\subsection{Sparse compositional metric learning with perturbation loss}
\label{sec:SCML_CR}

We start by briefly revisiting SCML~\cite{shi2014sparse}. The core idea is to represent the Mahalanobis distance as a non-negative combination of $K$ basis elements; that is,
\begin{equation*}
\footnotesize
\begin{split}
    \bm M = \sum_{k=1}^K w_k \bm b_k \bm b_k^T, \quad \bm w \geq 0,
\end{split}
\end{equation*}
where the basis set $\{\bm b_k \}_{k=1}^K$ is generated by using the Fisher discriminative analysis at several local regions. To learn a discriminative metric with good generalization ability, the learning objective comprises a margin-based hinge loss function and an $L_1$-norm regularization term as follows: 
\begin{equation*}
\footnotesize
\begin{split}
    \min_{\bm w} J_\text{SCML} = \frac{1}{|\mathcal{R}|} \sum_{\mathcal{R}} \left[1+ d^2_{\bm w}(\bm x_i, \bm x_j) - d^2_{\bm w}(\bm x_i, \bm x_l) \right]_+ + \eta \|\bm w\|_1,
\end{split}
\end{equation*}
where $\eta \geq 0$ controls the degree of sparsity. 

Now, we illustrate how to incorporate the perturbation loss into SCML and solve the associated optimization problem. The solution to the closest adversarial example and the form of perturbation loss remain the same as LMNN-PL; the learning of the Mahalanobis distance is replaced by learning the sparse coefficients, and the optimization problem is solved via the accelerated proximal gradient descent algorithm. 

First, we incorporate the proposed perturbation loss $J_\text{P}$ (Eq.~4) into the original objective function $J_\text{SCML}$:
\begin{equation}
\footnotesize
\begin{split}
    \min_{\bm M \in \mathbb{S}^p_+} J&= J_\text{SCML} + \lambda J_\text{P}.
\end{split}
\label{eq:obj_SCML}
\end{equation}
The squared adversarial margin of Eq.~\ref{eq:adv_margin} is now a function of $\bm w$:
\begin{equation*}
\scriptsize
\begin{split}
    &d^2_{\bm E}(\bm x_i, \bm x_{i,{\rm min}}) = \frac{[d^2_{\bm w} (\bm x_i, \bm x_l) - d^2_{\bm w} (\bm x_i, \bm x_j)]^2}{4 d^2_{\bm w^2}(\bm x_j, \bm x_l)} \\
    &d^2_{\bm w} (\bm x_i, \bm x_j) = (\bm x_i - \bm x_j)^T \big( \sum_{k=1}^K w_k \bm b_k \bm b_k^T \big) (\bm x_i - \bm x_j) \\
    &d^2_{\bm w^2}(\bm x_j, \bm x_l) = (\bm x_j - \bm x_l)^T \big(\sum_{k_1=1}^K \sum_{k_2=1}^K w_{k_1} w_{k_2} {\bm b}_{k_1} {\bm b}_{k_1}^T {\bm b}_{k_2} {\bm b}_{k_2}^T \big) (\bm x_j - \bm x_l).
\end{split}
\end{equation*}

Next, to optimize problem (\ref{eq:obj_SCML}), we apply the accelerated proximal gradient descent algorithm with a backtracking stepsize rule~\cite{Tibshirani-note}. The gradient of $J_\text{P}$ with respect to $\bm w$ is as follows: 
\begin{equation*}
\scriptsize
\begin{split}
    \frac{\partial J_\text{P}}{\partial w_k} =& \frac{1}{|\mathcal{R}|} \sum_\mathcal{R} \alpha_{ijl} \bigg\{ \frac{d^2_{\bm M}(\bm x_i, \bm x_l) - d^2_{\bm M}(\bm x_i, \bm x_j)}{2\left(d^2_{\bm M^2} (\bm x_j, \bm x_l)+\epsilon\right)} \tr \left(\bm b_k \bm b_k^T (\bm X_{ij} - \bm X_{il}) \right) \\
    +& \frac{[d^2_{\bm M}(\bm x_i, \bm x_l) - d^2_{\bm M}(\bm x_i, \bm x_j)]^2}{4\left(d^2_{\bm M^2} (\bm x_j, \bm x_l)+\epsilon\right)^2}\tr \left((\bm b_k \bm b_k^T \bm M + \bm M \bm b_k \bm b_k^T) \bm X_{jl}\right) \bigg\},
\end{split}
\end{equation*}
where $\epsilon$ is a small constant added to the denominator of $d^2_{\bm E}(\bm x_i, \bm x_{i,{\rm min}})$.

\subsection{Proxy neighborhood component analysis with perturbation loss}

Neighborhood component analysis is first proposed in~\cite{goldberger2004neighbourhood}, and in this paper, we build on one of its latest variants, ProxyNCA++~\cite{teh2020proxynca++}. ProxyNCA++ aims to maximize the proxy assignment probability as follows:
\begin{equation*}
\begin{split}
    J_\text{ProxyNCA++} = -\log\bigg(\frac{\exp\big(-d(\frac{\bm x_i}{\|\bm x_i\|_2},\frac{f(\bm x_i)}{\|f(\bm x_i)\|_2}) \cdot \frac{1}{T}\big)}{\sum_{f(\bm a) \in A}\exp\big(-d(\frac{\bm x_i}{\|\bm x_i\|_2},\frac{f(\bm a)}{\|f(\bm a)\|_2}) \cdot \frac{1}{T}\big)}\bigg),
\end{split}
\end{equation*}
where $f(\bm a)$ is a proxy function which maps each instance $\bm a$ to its class proxy, $A$ denotes the set of all proxies, and $T$ is the temperature scaling parameter. As the method is proposed in the context of deep metric learning, $\bm x_i$ is assumed to be the feature vector after the embedding network and $d$ is chosen as the Euclidean distance. In our work, we will use the original features and the Mahalanobis distance. 

There are two sets of learning parameters in ProxyNCA++, the Mahalanobis distance and the proxies. We optimize them alternatively, and in each step, we apply the gradient descent algorithm. The gradient of the ProxyNCA++ term is analogous to the classical NCA and can be found in~\cite{goldberger2004neighbourhood}. 

The perturbation loss has been revised slightly to include the adversarial margin from instances whose nearest proxy is from a different class:
\begin{equation*}
\begin{split}
    J_\text{P} = \frac{1}{n}\sum_{\bm x_i}
    \Big \{ 
    &[\tau^2 - \tilde{d}_{\bm E}^2(\bm x_i, \bm x_{i,{\rm min}})]_+ \mathds{1}[d^2_{\bm M}(\bm x_i, \bm a^-) > d^2_{\bm M}(\bm x_i, \bm a^+)] \\
    +&
    (\tau^2 + \tilde{d}_{\bm E}^2(\bm x_i, \bm x_{i,{\rm min}})) \mathds{1}[d^2_{\bm M}(\bm x_i, \bm a^-) \leq d^2_{\bm M}(\bm x_i, \bm a^+)]
    \Big\} ,
\end{split}
\end{equation*}
where $\bm a^+$ and $\bm a^-$ denote the nearest proxy to $\bm x_i$ from the same class and different class, respectively. The gradient of the perturbation loss with respect to $\bm M$ is given in Eq.~\ref{eq:grad}; the gradient of the perturbation loss with respect to the proxies is given as follows:
\begin{equation*}
\scriptsize
\begin{split}
\frac{\partial J_\text{P}}{\partial \bm a_i} &= \sum_{\bm x \in A_i^+}\Big\{ \frac{d^2_{\bm M}(\bm x, \bm a^-) - d^2_{\bm M}(\bm x, \bm a_i)}{(d^2_{\bm M^2} (\bm a_i, \bm a^-)+\epsilon)}\bm M (\bm a_i-\bm x) \\
&+ \frac{[d^2_{\bm M}(\bm x, \bm a^-) - d^2_{\bm M}(\bm x, \bm a_i)]^2}{2(d^2_{\bm M^2} (\bm a_i, \bm a^-)+\epsilon)^2}\bm M^2 (\bm a_i-\bm a^-)\Big\}\\
&+ \sum_{\bm x \in A_i^-}\Big\{- \frac{d^2_{\bm M}(\bm x, \bm a_i) - d^2_{\bm M}(\bm x, \bm a^+)}{(d^2_{\bm M^2} (\bm a^+, \bm a_i)+\epsilon)^2}\bm M (\bm a_i-\bm x) \\
&+ \frac{[d^2_{\bm M}(\bm x, \bm a_i) - d^2_{\bm M}(\bm x, \bm a^+)]^2}{2(d^2_{\bm M^2} (\bm a^+, \bm a_i)+\epsilon)^2}\bm M^2 (\bm a_i-\bm a^+)\Big\},
\end{split}
\end{equation*}
where $\bm x \in A_i^+$ is a set of instances whose nearest proxy is $\bm a_i$, and $\bm x$ and $\bm a_i$ have the same class label;  $\bm x \in A_i^-$ denotes a set of instances whose different-class nearest proxy is $\bm a_i$.

\section{Preliminaries and proof of generalization benefit}
\label{sec(app):generalization}

\subsection{Preliminaries}

\begin{defn}\cite{bellet2015robustness}
    An algorithm $\mathcal{A}$ is $(K,\epsilon(\cdot), \hat{n}(\cdot))$ \emph{pseudo-robust} for $K \in \mathbb{N}$, $\epsilon(\cdot): (\mathcal{Z} \times \mathcal{Z} \times \mathcal{Z})^n \rightarrow \mathbb{R}$ and $\hat{n}(\cdot): (\mathcal{Z} \times \mathcal{Z} \times \mathcal{Z})^n \rightarrow \{1, \ldots, n^3\}$ if $\mathcal{Z}=(\mathcal{X} \times \mathcal{Y})$ can be partitioned into $K$ disjoint sets, denoted by $\{C_k\}_{k=1}^K$, such that for all training samples $\bm s \in \mathcal{Z}^n$ drawn independently and identically distributed~(IID) from the probability distribution $\mathcal{P}$, there exists a subset of training triplets $\hat{t}_{\bm s} \subseteq t_{\bm s}$, with $|\hat{t}_{\bm s}|=\hat{n}(t_{\bm s})$, such that the following holds: $\forall (\bm s_1, \bm s_2, \bm s_3) \in \hat{t}_s$, $\forall \bm z_1, \bm z_2, \bm z_3 \in \mathcal{Z}$, $\forall i,j,l = 1, \ldots, K$, if $\bm s_1, \bm z_1 \in C_i$, $\bm s_2, \bm z_2 \in C_j$ and $\bm s_3, \bm z_3 \in C_l$, then
    \begin{equation*}
    \footnotesize
    \begin{split}
        |\ell(\mathcal{A}_{t_{\bm s}},\bm s_1, \bm s_2, \bm s_3) - \ell(\mathcal{A}_{t_{\bm s}},\bm z_1, \bm z_2, \bm z_3)| \leq \epsilon(t_{\bm s}) .
    \end{split}
    \end{equation*}
\end{defn}

\begin{thm}\cite{bellet2015robustness}
    If $\mathcal{A}$ is $(K,\epsilon(\cdot), \hat{n}(\cdot))$ pseudo-robust and the training triplets $t_{\bm s}$ come from a sample generated by $n$ IID draws from $\mathcal{P}$, then for any $\delta>0$, with probability at least $1-\delta$ we have:
    \begin{equation}
    \footnotesize
    \begin{split}
        &|\mathcal{L}(\mathcal{
        A}_{t_{\bm s}})-\ell_{emp}(\mathcal{
        A}_{t_{\bm s}})|  \\
        \leq& \frac{\hat{n}(t_{\bm s})}{n^3} \epsilon(t_{\bm s}) + B\Big(\frac{n^3 - \hat{n}(t_{\bm s})}{n^3} + 3\sqrt{\frac{2K \ln 2+2\ln 1/\delta}{n}}\Big),
    \end{split}
    \end{equation}
    where $\mathcal{L}$ denotes the expected error, $\ell_{emp}$ is the empirical error, and $B$ is a constant denoting the upper bound of the loss function $\ell$. $|\mathcal{L}-\ell_{emp}|$ is termed the generalization gap.
\end{thm}

\begin{defn}\cite{wainwright_2019}
    A \emph{$\delta$-cover} of a set $\Theta$ with respect to a metric $\rho$ is a set $\{\theta^1, \ldots, \theta^N\} \subset \Theta$ such that for each $\theta \in \Theta$, there exists some $i\in \{ 1,\ldots, N\}$ such that $\rho(\theta,\theta^i) \leq \delta$. The \emph{$\delta$-covering} number $N(\delta, \Theta,\rho)$ is the cardinality of the smallest $\delta$-cover.
\end{defn}

\subsection{Theorem and proof}

\begin{thm}
    Let $\bm M^\ast$ be the optimal solution to Eq.~\ref{eq:obj}. Then for any $\delta>0$, with probability at least $1-\delta$ we have:
    \begin{equation*}
    \footnotesize
    \begin{split}
        &|\mathcal{L}(\bm M^\ast)-\ell_{emp}(\bm M^\ast)| \\
        \leq& \frac{\hat{n}(t_{\bm s})}{n^3} + B\Big(\frac{n^3 - \hat{n}(t_{\bm s})}{n^3} + 3\sqrt{\frac{2K \ln 2+2\ln 1/\delta}{n}}\Big),
    \end{split}
    \end{equation*}
    where $\hat{n}(t_{\bm s})$ denotes the number of triplets whose adversarial margins are larger than $\tau$, $B$ is a constant denoting the upper bound of the loss function (i.e. Eq.~\ref{eq:obj}), and $K$ denotes the number of disjoint sets that partition the input-label space and equals to $|\mathcal{Y}|(1+\frac{2}{\tau})^p$.
\end{thm}
\begin{proof}
    After embedding the perturbation loss, learning algorithms that minimize the classical triplet loss, i.e. $\left[1+ d^2_{\bm M}(\bm x_i, \bm x_j) - d^2_{\bm M}(\bm x_i, \bm x_l) \right]_+ \cdot \mathds{1}_{\{y_i=y_j \neq y_l\}}$, are $(|\mathcal{Y}|(1+\frac{2}{\tau})^p,1,\hat{n}(\cdot;\tau))$ pseudo-robust. $\epsilon=1$ since, by definition of adversarial margin, any $\bm x$ that falls into the Euclidean ball with center $\bm x_i$ and a radius of the desired margin $\tau$ will satisfy $d^2_{\bm M}(\bm x, \bm x_l) > d^2_{\bm M}(\bm x, \bm x_j)$. Therefore, any change in the triplet loss is bounded by 1. The value of $K$ can be determined via the covering number~\cite{bellet2015robustness}. The instance space $\mathcal{X}$ can be partitioned by using the covering number $N(\tau,\mathcal{X},\|\cdot\|_2)$. By normalizing all instances to have unit $L_2$-norm, we obtain a finite covering number as $N \leq (1+\frac{2}{\tau})^p$~\cite{wainwright_2019}. The label space $\mathcal{Y}$ can be partitioned into $|\mathcal{Y}|$ sets. Therefore, the number of disjoint sets, i.e. $K$, is always smaller than $|\mathcal{Y}|(1+\frac{2}{\tau})^p$. 
\end{proof}

\section{Experimental setup and additional results}
\label{sec(app):exp}

\subsection{Datasets}
\label{subsec(app):data}

Table~\ref{tab:data} lists information on sample size, feature dimension and class information. The class distribution of datasets Ecoli, Yeast and APS Failure is highly imbalanced and their imbalanced ratios (IR) are also listed. 

\begin{table}[htbp]
\caption{Characteristics of the datasets.}
\label{tab:data}
\centering
\small
\setlength\tabcolsep{1pt}
\begin{tabular}{lccccc}
\toprule
Dataset & \# instances & \# features & \# classes & \# rounds\\
 & (training,test,valid.) & (reduced dimensions)  \\
\midrule
\multicolumn{5}{c}{UCI data}\\
Australian & 690 & 14 & 2 & 20\\
Breast cancer & 683  & 9 & 2 & 20\\
Ecoli & 336 & 7 & 2 (IR:10.59) & 20\\
(0-1-4-7 vs 2-3-5-6)\\
Fourclass & 862 & 2 & 2 & 20\\
Haberman & 306 & 3 & 2 & 20\\
Iris & 150 & 4 & 3 & 20\\
Segment & 2310 & 19 & 7 & 20\\
Sonar & 208 & 60 & 2 & 20\\
Voting & 435 & 16 & 2 & 20\\
WDBC & 569 & 30 & 2 & 20\\
Wine & 178 & 13 & 3 & 20\\
Yeast (CYT vs POX) & 483 & 8 & 2 (IR:23.15) & 20\\
Credit (Taiwan) & 30000 & 23 & 2 & 1\\
&(10k,10k,10k)\\
\midrule
\multicolumn{5}{c}{High-dimensional data}\\
Isolet & 7797 & 617 & 26 & 4\\
& (1560,1558) & (170) \\
MNIST & 4000 & 784 & 10 & 1\\
& (2000,2000) & (141) \\
APS Failure & 76000 & 171 & 2 (IR:58.90) & 1\\
& (40k,20k,16k) & (79) \\
\bottomrule
    \end{tabular}
\end{table}

\subsection{Experimental setting}
\label{subsec(app):exp_setting}

\begin{figure*}[t]
\centering
\subfloat[isotropic Gaussian noise]{\makebox[.3\linewidth][c]{\includegraphics[width=.3\linewidth]{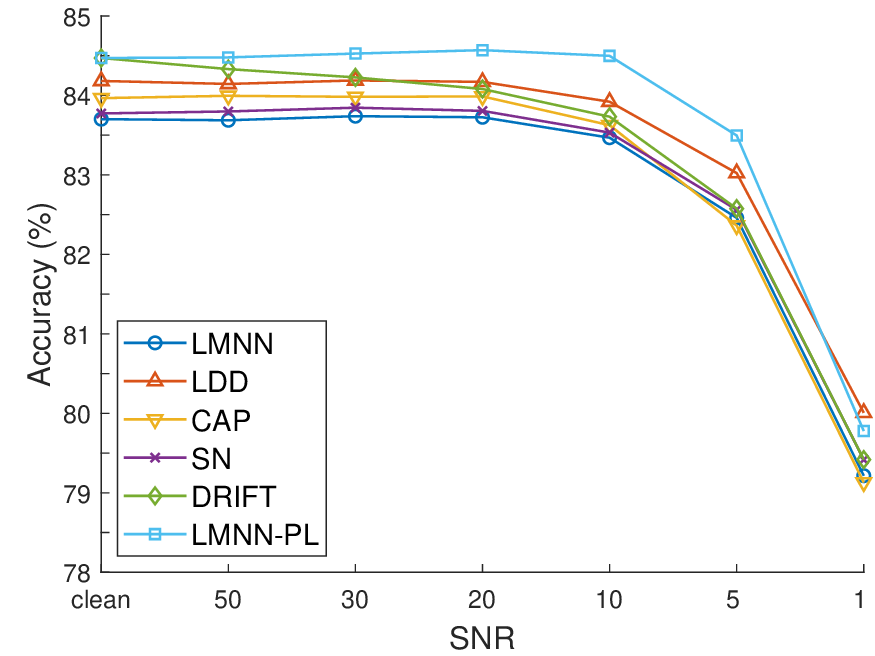}}
\label{fig:AU_sphere}}\hspace{.1em}
\subfloat[anisotropic Gaussian noise]{\makebox[.3\linewidth][c]{\includegraphics[width=.3\linewidth]{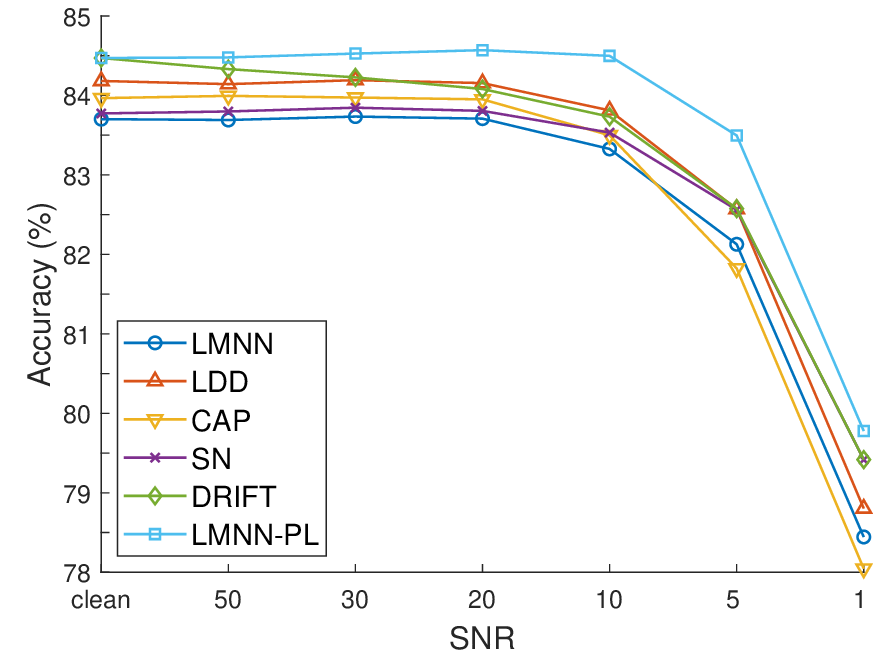}}
\label{fig:AU_diagonal}}\hspace{.1em}
\subfloat[adversarial perturbations]{\makebox[.3\linewidth][c]{\includegraphics[width=.3\linewidth]{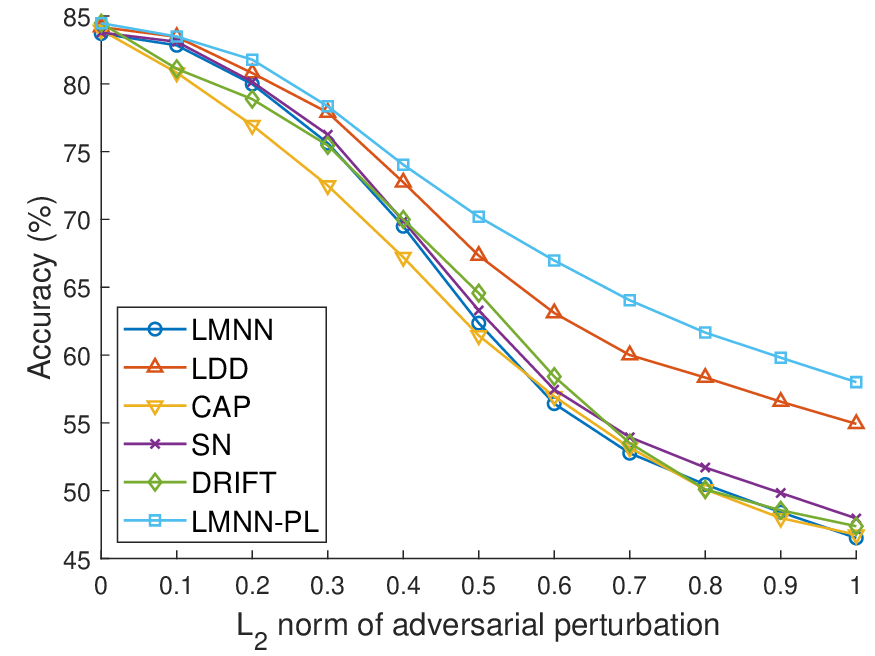}}
\label{fig:AU_CR}}
\caption{Robustness of LMNN-based methods on the Australian dataset with different noise types and intensities.}
\end{figure*}

\subsubsection*{Hyperparameter tuning of compared methods}

LMNN, SCML, and DRIFT are implemented by using the official codes provided by the authors; all parameters are set as default apart from the trade-off parameter. For LMNN, the trade-off parameter $\mu$ is chosen from $\{0.1,0.2,\ldots,0.9\}$. For SCML, the weight of the regularization term $\eta$ is chosen from $\{10^{-5}, 10^{-4}, \ldots, 10^3\}$, and the number of bases is set as 200, 400 and 1000 for UCI datasets whose sample size is smaller than 500, larger than 500, and large-scale/high-dimensional datasets (i.e. Credit, Isolet, and MNIST), respectively. For LDD, the regularizer weight is chosen from $\{10^{-6},10^{-5}, \ldots, 10^2\}$. For CAP, the regularizer weight is chosen from $\{10^{-3},10^{-2}, \ldots, 10\}$, and the rank of $\bm M$ is chosen from 10 values equally spaced between 1 and $p$. For PNCA, the temperature scaling parameter is chosen from $\{10^{-2},10^{-1}, \ldots, 10^2\}$, and the number of proxies per class is chosen from $[1,2,3,5,10]$ for all UCI datasets except $[1,3,5,10,20]$ for Credit. For DRIFT, we search the grid suggested by the authors. Trade-off parameters are tuned via five-fold cross-validation on the training data or on a separate validation set as listed in Table~\ref{tab:data}. For all these methods, triplet constraints are generated from 3 target neighbors and 10 nearest impostors, calculated under the Euclidean distance.

\subsubsection*{Experimental setting of LMNN-PL}
$\bm M$ is initialized as the identity matrix. The learning rate $\gamma$ is initialized to 1. Following~\cite{weinberger2009distance}'s work, $\gamma$ is increased by $1\%$ if the loss function decreases and decreased by $50\%$ otherwise. The training stops if the relative change in the objective function is smaller than the threshold of $10^{-7}$ or reaches the maximum number of iterations of 1000.

\subsubsection*{Experimental setting of SCML-PL}
SCML-PL is tuned in the same manner as LMNN-PL via random search; the range of $\eta$ and the number of bases are same as SCML, and the ranges of $\tau$ and $\lambda$ are same as LMNN-PL. The method is optimised via the accelerated proximal gradient descent algorithm with a backtracking stepsize rule~\cite{Tibshirani-note}. The initial learning rate is set as 1 and the shrinkage factor is set as 0.8. $\bm w$ is initialized as the unit vector.

\subsubsection*{Experimental setting of PNCA-PL}
Ranges of hyperparameters are same as PNCA and LMNN-PL, except that $\lambda$ is chosen from $\{10^{-2},10^{-1},\ldots,10^2\}$. In addition, for training stability, we update proxies only in the first 200 iterations and fix them afterwards.

\subsection{Additional experimental results}
\label{sec:additional_exp}

\begin{sidewaystable}
\caption{Classification accuracy (or G-means indicated by an asterisk) of 3NN on datasets contaminated with anisotropic Gaussian noise (SNR$=$5~dB).}
\label{tab:noise_elliptical}
\small
\centering
\setlength\tabcolsep{2pt}
\begin{tabular}{lccc cccc cccc ccc}
\toprule
&&& \multicolumn{6}{c}{LMNN-based}&& \multicolumn{2}{c}{SCML-based}&& \multicolumn{2}{c}{PNCA-based}\\
\cmidrule{4-9} \cmidrule{11-12} \cmidrule{14-15}
Dataset & Euclidean &  & LMNN & LDD & CAP & SN & DRIFT & LMNN-PL &  & SCML & SCML-PL &  & PNCA & PNCA-PL\\
\midrule
Australian & 81.88$\pm$1.72 &  & 82.13$\pm$1.52 & \underline{82.57$\pm$1.55} & 81.82$\pm$1.52 & 82.20$\pm$1.54 & 81.97$\pm$1.53 & \textbf{82.90$\pm$1.53} &  & 82.59$\pm$1.70 & \textbf{82.84$\pm$1.64} &  & 83.21$\pm$1.69 & \textbf{83.68$\pm$1.81}
\\
Breast cancer & 96.76$\pm$1.06 &  & 96.24$\pm$1.06 & \underline{96.66$\pm$1.07} & 96.27$\pm$1.03 & 96.27$\pm$1.07 & 96.61$\pm$0.97 & \textbf{96.69$\pm$1.07} &  & 96.34$\pm$1.03 & \textbf{96.63$\pm$1.03} &  & 96.05$\pm$1.39 & \textbf{96.74$\pm$1.13}
\\
Ecoli* & 78.28$\pm$7.53 &  & 75.13$\pm$7.58 & \underline{76.97$\pm$7.64} & 72.64$\pm$9.06 & 75.27$\pm$7.60 & 76.18$\pm$7.65 & \textbf{78.58$\pm$7.53} &  & 75.02$\pm$6.89 & \textbf{77.14$\pm$7.97} &  & 90.91$\pm$3.17 & \textbf{91.40$\pm$2.55}
\\
Fourclass & 69.20$\pm$1.11 &  & 67.74$\pm$1.25 & 68.84$\pm$1.14 & 67.84$\pm$1.19 & 68.61$\pm$1.26 & \textbf{69.13$\pm$1.02} & \underline{69.04$\pm$1.11} &  & 68.22$\pm$1.10 & \textbf{68.96$\pm$1.11} &  & \textbf{70.54$\pm$2.14} & 69.49$\pm$4.62
\\
Haberman & 70.30$\pm$1.89 &  & 70.21$\pm$1.84 & \textbf{70.25$\pm$1.82} & 69.39$\pm$2.06 & 70.23$\pm$1.85 & 69.31$\pm$2.50 & \textbf{70.25$\pm$1.90} &  & 69.98$\pm$1.64 & \textbf{70.24$\pm$1.85} &  & 74.72$\pm$3.15 & \textbf{74.76$\pm$3.09}
\\
Iris & 79.84$\pm$3.28 &  & 78.75$\pm$2.96 & \underline{79.04$\pm$3.17} & 77.90$\pm$3.31 & 78.84$\pm$3.08 & 78.57$\pm$3.09 & \textbf{79.20$\pm$3.08} &  & 78.32$\pm$3.60 & \textbf{79.18$\pm$3.13} &  & 78.13$\pm$4.55 & \textbf{78.34$\pm$3.91}
\\
Segment & 86.27$\pm$0.70 &  & 79.03$\pm$3.37 & \underline{83.49$\pm$1.17} & 82.77$\pm$2.49 & 80.04$\pm$3.25 & \textbf{83.88$\pm$1.33} & 82.13$\pm$2.70 &  & 61.28$\pm$9.78 & \textbf{62.86$\pm$8.76} &  & 76.30$\pm$2.74 & \textbf{78.28$\pm$3.32}
\\
Sonar & 83.50$\pm$3.19 &  & 83.54$\pm$4.30 & \textbf{86.18$\pm$2.93} & \underline{85.44$\pm$2.79} & 83.50$\pm$4.32 & 84.65$\pm$3.30 & 84.99$\pm$3.13 &  & 76.91$\pm$4.32 & \textbf{79.49$\pm$3.80} &  & 80.71$\pm$4.35 & \textbf{81.76$\pm$3.43}
\\
Voting & 93.19$\pm$1.14 &  & 94.01$\pm$1.00 & \underline{94.24$\pm$1.13} & 94.37$\pm$1.17 & 94.05$\pm$1.01 & 93.94$\pm$1.12 & \textbf{94.64$\pm$1.21} &  & 93.99$\pm$1.15 & \textbf{94.65$\pm$1.09} &  & 92.61$\pm$1.64 & \textbf{93.45$\pm$1.77}
\\
WDBC & 95.89$\pm$1.31 &  & 92.01$\pm$1.65 & \textbf{96.30$\pm$0.94} & \underline{96.14$\pm$1.11} & 92.67$\pm$1.69 & 96.02$\pm$0.88 & 96.07$\pm$0.89 &  & 95.75$\pm$1.29 & \textbf{96.22$\pm$1.14} &  & 96.02$\pm$1.51 & \textbf{96.24$\pm$1.14}
\\
Wine & 94.13$\pm$1.47 &  & 93.27$\pm$1.62 & 93.97$\pm$1.38 & 93.87$\pm$1.49 & 93.38$\pm$1.68 & \textbf{94.55$\pm$1.15} & \underline{94.44$\pm$1.21} &  & 93.92$\pm$1.55 & \textbf{94.52$\pm$1.33} &  & \textbf{94.08$\pm$1.93} & 93.37$\pm$1.49
\\
Yeast* & 69.91$\pm$10.46 &  & 59.00$\pm$8.78 & 68.63$\pm$10.66 & 65.74$\pm$10.26 & 59.58$\pm$8.98 & \underline{70.15$\pm$10.56} & \textbf{70.23$\pm$10.48} &  & 56.43$\pm$8.74 & \textbf{60.32$\pm$9.07} &  & 61.84$\pm$18.45 & \textbf{63.70$\pm$18.69}
\\
Credit & 76.01 &  & 76.07 & \underline{76.22} & 74.52 & \textbf{76.29} & 76.12 & 76.06 &  & \textbf{75.73} & \textbf{75.73} &  & 78.89 & \textbf{79.73}\\
\midrule
\# outperform & - &  & 12 & 9 & 10 & 12 & 9 & - &  & 13 & - &  & 11 & -\\
\bottomrule
\end{tabular}

\vspace{.3in}
\caption{Classification accuracy (or G-means indicated by an asterisk) of 3NN on datasets contaminated with adversarial perturbations ($L_2$-norm$=$0.2).}
\label{tab:noise_adv}
\small
\centering
\setlength\tabcolsep{2pt}
\begin{tabular}{lccc cccc cccc ccc}
\toprule
&&& \multicolumn{6}{c}{LMNN-based}&& \multicolumn{2}{c}{SCML-based}&& \multicolumn{2}{c}{PNCA-based}\\
\cmidrule{4-9} \cmidrule{11-12} \cmidrule{14-15}
Dataset & Euclidean &  & LMNN & LDD & CAP & SN & DRIFT & LMNN-PL &  & SCML & SCML-PL &  & PNCA & PNCA-PL\\
\midrule
Australian & 79.09$\pm$2.57 &  & 79.98$\pm$2.95 & \underline{80.79$\pm$2.72} & 76.95$\pm$3.86 & 80.14$\pm$2.78 & 78.87$\pm$3.13 & \textbf{81.78$\pm$2.53} &  & 79.64$\pm$3.11 & \textbf{81.08$\pm$2.58} &  & 83.89$\pm$2.91 & \textbf{84.35$\pm$3.04}
\\
Breast cancer & 96.29$\pm$1.57 &  & 95.44$\pm$1.69 & \textbf{96.00$\pm$1.84} & 95.68$\pm$1.44 & 95.46$\pm$1.77 & 95.85$\pm$1.67 & \textbf{96.00$\pm$1.53} &  & 95.59$\pm$1.60 & \textbf{95.71$\pm$1.66} &  & 96.37$\pm$1.91 & \textbf{96.66$\pm$1.67}
\\
Ecoli* & 80.60$\pm$11.55 &  & 72.82$\pm$16.67 & \textbf{80.70$\pm$11.42} & 68.78$\pm$21.39 & 73.84$\pm$16.43 & 78.59$\pm$10.55 & \underline{79.97$\pm$13.37} &  & 74.80$\pm$14.62 & \textbf{75.20$\pm$13.99} &  & 71.58$\pm$20.12 & \textbf{79.07$\pm$9.35}
\\
Fourclass & 66.27$\pm$2.64 &  & 60.73$\pm$4.97 & \textbf{66.53$\pm$2.34} & 64.34$\pm$3.18 & 64.05$\pm$4.44 & 66.22$\pm$2.66 & \underline{66.35$\pm$3.00} &  & 64.90$\pm$3.18 & \textbf{66.49$\pm$2.89} &  & \textbf{72.80$\pm$3.32} & 71.29$\pm$5.21
\\
Haberman & 57.58$\pm$4.26 &  & 57.25$\pm$5.28 & 57.19$\pm$4.67 & \underline{57.53$\pm$5.07} & 57.19$\pm$5.05 & \textbf{59.16$\pm$6.17} & 57.30$\pm$4.55 &  & 56.40$\pm$5.37 & \textbf{57.02$\pm$3.75} &  & 74.94$\pm$4.39 & \textbf{75.45$\pm$4.38}
\\
Iris & 78.67$\pm$5.37 &  & 76.33$\pm$6.70 & 76.67$\pm$6.22 & \textbf{76.78$\pm$5.42} & 76.11$\pm$7.06 & 76.44$\pm$6.93 & \textbf{76.78$\pm$6.39} &  & 74.00$\pm$6.99 & \textbf{76.67$\pm$6.26} &  & \textbf{78.22$\pm$8.14} & 77.67$\pm$7.33
\\
Segment & 76.10$\pm$1.70 &  & 56.80$\pm$6.21 & \underline{67.86$\pm$3.12} & 67.37$\pm$6.19 & 59.21$\pm$6.54 & \textbf{71.31$\pm$2.83} & 65.53$\pm$5.17 &  & 29.49$\pm$9.00 & \textbf{30.56$\pm$12.95} &  & 68.71$\pm$4.31 & \textbf{71.61$\pm$4.66}
\\
Sonar & 71.75$\pm$6.11 &  & 42.38$\pm$16.87 & \underline{64.13$\pm$6.36} & 58.33$\pm$6.89 & 41.67$\pm$16.92 & \textbf{64.44$\pm$5.46} & 56.03$\pm$8.43 &  & 28.17$\pm$11.22 & \textbf{40.24$\pm$11.81} &  & 60.40$\pm$5.93 & \textbf{61.90$\pm$11.01}
\\
Voting & 90.88$\pm$2.06 &  & 89.54$\pm$2.97 & 91.11$\pm$2.62 & 91.03$\pm$2.70 & 89.47$\pm$3.05 & \underline{91.22$\pm$2.26} & \textbf{92.71$\pm$1.99} &  & 90.08$\pm$2.46 & \textbf{92.25$\pm$2.92} &  & 87.79$\pm$3.70 & \textbf{89.89$\pm$6.93}
\\
WDBC & 93.57$\pm$1.89 &  & 59.27$\pm$12.73 & \textbf{91.05$\pm$2.08} & 90.56$\pm$3.02 & 65.03$\pm$12.00 & \underline{90.91$\pm$2.40} & 90.41$\pm$3.58 &  & 87.51$\pm$4.15 & \textbf{90.73$\pm$2.97} &  & 90.85$\pm$3.24 & \textbf{91.64$\pm$2.64}
\\
Wine & 90.28$\pm$3.23 &  & 84.07$\pm$6.02 & \textbf{90.09$\pm$3.08} & 86.94$\pm$3.53 & 84.63$\pm$6.16 & 87.59$\pm$3.40 & \underline{88.43$\pm$4.07} &  & 86.02$\pm$5.39 & \textbf{88.98$\pm$4.55} &  & \textbf{88.70$\pm$3.75} & 86.20$\pm$4.55
\\
Yeast* & 70.21$\pm$10.47 &  & 53.42$\pm$26.38 & \textbf{70.19$\pm$10.50} & 68.15$\pm$12.59 & 52.15$\pm$25.54 & 70.14$\pm$10.49 & \textbf{70.19$\pm$10.49} &  & 59.83$\pm$13.44 & \textbf{62.61$\pm$18.96} &  & 61.65$\pm$19.36 & \textbf{66.65$\pm$19.05}
\\
Credit & 67.22 &  & 67.14 & 67.21 & 65.02 & \textbf{67.29} & \underline{67.23} & 66.96 &  & 66.03 & \textbf{66.44} &  & \textbf{80.60} & 80.27\\
\midrule
 & - &  & 12 & 6 & 9 & 12 & 8 & - &  & 13 & - &  & 9 & -\\
\bottomrule
\end{tabular}
\end{sidewaystable}

In addition to the isotropic Gaussian noise presented in the main text, we test the robustness performance against anistropic Gaussian noise and adversarial perturbations. Anistropic Gaussian noise is generated from a zero-mean Gaussian with a diagonal covariance matrix and unequal variances estimated from the training data; its noise intensity is determined via SNR. Adversarial perturbations are computed according to Eq.~\ref{eq:x_min_def} with the nearest target neighbor and impostor found from using the learned distance metric; the magnitude of perturbations is controlled via $L_2$ norm.

To start with, we conduct an in-depth experiment on the Australian dataset by altering the noise intensity. Fig.~\ref{fig:AU_sphere} plots the classification accuracy of LMNN-based methods under different levels of isotropic Gaussian noise (equal variances). When the noise intensity is low, the performance of LMNN and LMNN-PL remain stable. When the noise intensity increases to the SNR of 10~dB or 5~dB, the performances of both method degrade. Owing to the enlarged adversarial margin, the influence on LMNN-PL is slightly smaller than that on LMNN. When the SNR equals 1~dB, the performance gain from using LMNN-PL becomes smaller. This result is reasonable as the desired margin $\tau$ is selected according to the criterion of classification accuracy and hence may be too small to withstand a high level of noise. LMNN-PL surpasses all other LMNN-based methods until the noise intensity is very large. Fig.~\ref{fig:AU_diagonal} plots the accuracy under anisotropic Gaussian noise (unequal variances). Compared with the case of isotropic Gaussian noise, the degradation of all methods is more pronounced in this case, but the pattern remains similar. Fig.~\ref{fig:AU_CR} presents the results under adversarial perturbations. The proposed method achieves the highest accuracy over the range of perturbation size. The method LDD is also quite robust to adversarial perturbations. This should not be surprising as it encourages learning a metric close to the Euclidean distance, and the Euclidean distance is less sensitive to perturbation than the discriminative Mahalanobis distance.

Tables~\ref{tab:noise_elliptical} and~\ref{tab:noise_adv} list the classification accuracy or G-means for all datasets contaminated by the anistropic Gaussian noise with the SNR of 5~dB and the adversarial perturbations with the perturbation size of 0.2, respectively. Comparing the case of adversarial perturbations with that of Gaussian-type noises, we see three clear differences. First, LMNN performs much worse than the Euclidean distance in the presence of adversarial perturbations, especially on datasets with a large number of features relative to the sample size such as Sonar and WDBC. A potential reason is that the learned metric will stretch the distances in a few directions for discriminability, and hence adding perturbations in these directions is very likely to change the decision of $k$NN. Secondly, robustness to adversarial perturbations differs markedly across the types of methods. SCML-based deteriorates drastically on some datasets while PNCA-based methods are much more robust. Thirdly, the proposed method is effective in safeguarding the baseline methods against adversarial perturbations on most of the datasets.  

Table~\ref{tab:HD_additional} is a supplement to Table~\ref{tab:HD} of main text. It reports the performance of Euclidean, AML, LDD and DRIFT on high-dimensional datasets. 

\begin{table}[tbp]
\caption{Generalization and robustness of metric learning methods on high-dimensional datasets (additional results).}
\label{tab:HD_additional}
\small
\centering
\setlength\tabcolsep{2.5pt}
\begin{threeparttable}
\begin{tabular}{lcccccc}
\toprule
\multicolumn{7}{c}{Isolet}\\
\midrule
Method & Clean & IG,SNR=20 & IG,SNR=5 & AG,SNR=20 & AG,SNR=5 & Margin \\
\midrule
Euclidean & 84.16$\pm$4.09 & 83.93$\pm$4.30 & 82.73$\pm$3.70 & 83.93$\pm$4.30 & 83.18$\pm$3.59 & 0.1009\\
AML & 86.75$\pm$3.16 & 86.59$\pm$3.49 & 85.97$\pm$3.69 & 86.69$\pm$3.59 & 86.24$\pm$3.82 & 0.1261\\
LDD & 90.91$\pm$3.90 & 90.81$\pm$4.12 & 87.97$\pm$3.83 & 90.75$\pm$4.12 & 89.13$\pm$4.05 & 0.1333\\
\bottomrule
\toprule
\multicolumn{7}{c}{MNIST}\\
\midrule
Method & Clean & IG,SNR=20 & IG,SNR=5 & AG,SNR=20 & AG,SNR=5 & Margin\\
\midrule
Euclidean & 88.70 & 88.90 & 88.25 & 89.05 & 88.05 & 0.2091\\
AML & 89.25 & 88.70 & 88.85 & 89.30 & 89.20 & 0.2142\\
LDD & 90.85 & 90.85 & 87.90 & 90.95 & 90.30 & 0.2232\\
DRIFT & 90.85 & 90.75 & 87.45 & 90.65 & 89.45 & 0.2054\\
\bottomrule
\end{tabular}
\smallskip
\textit{DRIFT is unable to learn a metric on Isolet and hence is not reported.}
\end{threeparttable}
\end{table}

\subsection{Convergence analysis}
The objective function of the proposed perturbation loss (Eq.~\ref{eq:perturbation_loss}) includes several indicator functions. As a consequence, we cannot guarantee that the optimization algorithm will converge to a global or local optimum. Empirically, we decay the learning rate as training progresses and stop training when the change in the objective function is small or if it reaches the maximum iteration. Figure~\ref{fig:MNIST_conv} shows the classification accuracy of the proposed LMNN-PL on the MNIST dataset along training. The test accuracy increases in the initial stage, oscillates afterwards, and stabilizes after around 500 iterations. 

\begin{figure}[htbp]
    \centering
    \includegraphics[width=.5\linewidth]{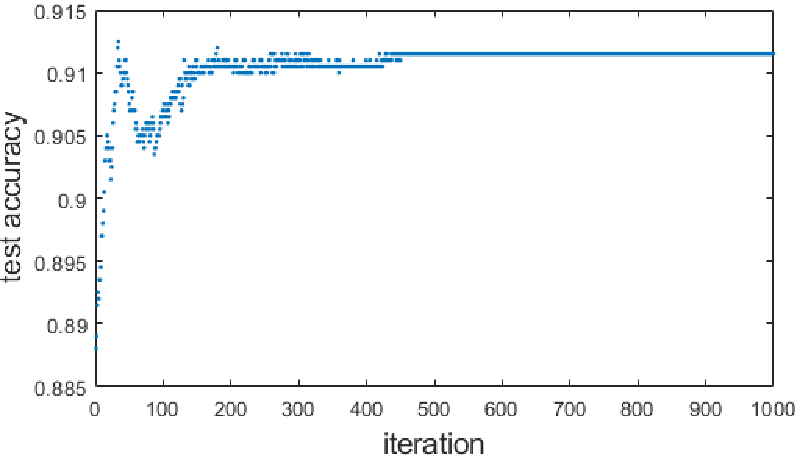}
    \caption{Convergence curve of LMNN-PL on MNIST.}
    \label{fig:MNIST_conv}
    \vspace{3.5in}
\end{figure}

\end{document}